\documentclass{article}
\usepackage[a4paper, total={6in, 8in}]{geometry}
\usepackage{amsmath,amsfonts}
\usepackage{algorithmic}
\usepackage{algorithm}
\usepackage{array}
\usepackage[caption=false,font=normalsize,labelfont=sf,textfont=sf]{subfig}
\usepackage{textcomp}
\usepackage{stfloats}
\usepackage{url}
\usepackage{verbatim}
\usepackage{graphicx}
\usepackage{cite}
\usepackage{xcolor}
\usepackage{multirow}
\usepackage{adjustbox}
\usepackage{hyperref}

\begin{document}

\title{A robust approach for deep neural networks in presence of label noise: relabelling and filtering instances during training}

\author{Anabel Gómez-Ríos\textsuperscript{a}, Julián Luengo\textsuperscript{a}, Francisco Herrera\textsuperscript{a}\\
\textsuperscript{a} Department of Computer Science and Artificial Intelligence,\\ Andalusian Research Institute in Data Science and Computational Intelligence\\(DaSCI), University of Granada, 18071, Granada, Spain}
\date{}



\maketitle

\begin{abstract}
Deep learning has outperformed other machine learning algorithms in a variety of tasks, and as a result, it is widely used. However, like other machine learning algorithms, deep learning, and convolutional neural networks (CNNs) in particular, perform worse when the data sets present label noise. Therefore, it is important to develop algorithms that help the training of deep networks and their generalization to noise-free test sets. In this paper, we propose a robust training strategy against label noise, called RAFNI, that can be used with any CNN. This algorithm filters and relabels instances of the training set based on the predictions and their probabilities made by the backbone neural network during the training process. That way, this algorithm improves the generalization ability of the CNN on its own. RAFNI consists of three mechanisms: two mechanisms that filter instances and one mechanism that relabels instances. In addition, it does not suppose that the noise rate is known nor does it need to be estimated. We evaluated our algorithm using different data sets of several sizes and characteristics. We also compared it with state-of-the-art models using the CIFAR10 and CIFAR100 benchmarks under different types and rates of label noise and found that RAFNI achieves better results in most cases.
\end{abstract}


\section{Introduction}
\label{sec:introduction}
Over the last few years, deep learning and Convolutional Neural Networks (CNNs) in particular have become progressively popular as they have been used in a variety of applications, especially in computer vision, and outperform other models \cite{krizhevsky2012imagenet, GOMEZRIOS2019315, olmos2018automatic}. Generally speaking, the networks that have been used in these types of applications have become deeper and deeper over the years, due to their high performance.

A common problem when dealing with real-world data sets in the context of supervised classification is label noise. The term `label noise' refers to when some instances in the data set have erroneous labels, thus misleading the training of machine learning algorithms \cite{song2020learning}. This type of noise can be present in the data set because it was labelled automatically using text labels from the Internet, or because not enough experts were available to label an entire data set. In either case, the rate of label noise can vary and can increase to large values \cite{xiao2015learning, lee2017cleannet}. As a result, label noise has been extensively studied when using classical machine learning algorithms \cite{6685834}. The two most used and studied types of label noise are symmetric and asymmetric noise. In symmetric noise, also called uniform noise, the labels are corrupted randomly and equally in all classes, independently of their true class. In asymmetric noise, the corruptions are dependent on the true class of the instances, but not on the instances themselves. This implies that the corruptions in the labels can be made so that the instances in one specific class are labelled as another specific class. Subsequently, asymmetric noise is more realistic than symmetric noise. In \cite{6685834}, the authors made a taxonomy of label noise in classification, where symmetric noise is called Noisy Completely at Random (NCAR) and asymmetric noise is called Noisy at Random (NAR). There is another type of noise, called Noisy Not at Random (NNAR), which is the most realistic, where the corruptions are dependent on the true class of the instances and the instances themselves.

The use of increasingly deeper networks implies the need for larger data sets to adequately train them. This fact has caused researchers to investigate ways to overcome the lack of data. One possible solution is to create larger data sets by labelling them automatically instead of relying on experts, which is usually the only labelling solution for very large data sets \cite{xiao2015learning, li2017webvision, song2019selfie}. Another solution is to use techniques such as transfer learning and data augmentation. Transfer learning allows the network to start from a pre-trained state: instead of starting the training from scratch for every problem, the network is already pre-trained on another data set, usually larger and related to the new one in some way. As a consequence, transfer learning speeds up the training process. On the other hand, data augmentation artificially increases the size of the training data by introducing transformations of the original images, such as rotations, changes in lightning, cropping, flipping, etc. But, if the original training set presents label noise, the use of data augmentation can aggravate it, thus becoming another source of label noise.

In the specific case of deep learning, label noise has been proven to harm generalization when training deep neural networks \cite{zhang2021understanding}, and thus there has been an increasing amount of studies trying to improve the behaviour of deep neural networks as much as possible in presence of label noise \cite{patrini2017making, song2019selfie, song2020learning, ma2018dimensionality, jiang2018mentornet}. Label noise appears mostly in real-world data sets, where the noise rate is not known. However, to test the performance of the proposed models, we need a controlled environment where the noise is artificially introduced in some noise-free data sets using different noise rates. Two of the most used data sets for this are CIFAR10 and CIFAR100 \cite{wang2018iterative, 10.5555/3326943.3327112, jindal2016learning, patrini2017making}. Though it is necessary to test the proposals in large data sets like CIFAR or MNIST, it is also important to analyse them in other scenarios, like with small data sets. Small data sets are also common in real-world problems where it is not possible to collect more data. The majority of the current proposals lack this scenario.

To make the training of deep neural networks robust to all types of label noise, the analysis of incorrect classified instances for filtering and relabelling is the usual way to proceed. In this paper, we consider this hypothesis to deal with this problem, we consider our algorithm can identify and handle the noisy instances.

We propose an algorithm that, during the training process, relabels or filters the instances that it considers noisy using the predictions made by the backbone network. The backbone network is the deep neural network chosen to classify the data set (for instance, ResNet50), which will be trained using backpropagation as usual. Thus, we called the algorithm the Relabelling And Filtering Noisy Instances (RAFNI) algorithm. We have made the algorithm publicly available. As opposed to some of the previous proposals for this task, we do not suppose that the noise rate is known nor do we estimate it. Instead, the RAFNI algorithm uses the noisy training set and progressively cleans it during the training process by using the loss value of each instance and the probability of each instance belonging to each class. These values are given by the backbone network at each epoch of the training process. The algorithm is composed of two filtering mechanisms to remove noisy instances from the training set, and one relabelling mechanism that is used to change the class of some instances to their original (clean) class.

We evaluated our proposal with a variety of data sets, including small and large data sets, and under different types of label noise. We also use CIFAR10 and CIFAR100 as benchmarks to compare our proposal with other state-of-the-art models, since these data sets are two of the most used in other studies.

The rest of the paper is organized as follows. In Section~\ref{sec:background}, we give a background on works that propose strategies to help neural networks learn with label noise, with special mention to the ones we compare our algorithm to. In Section~\ref{sec:proposal}, we provide a detailed description of the RAFNI algorithm and its hyperparameters. Section~\ref{sec:experimentalFramework} details the experimental framework, including the data sets, the types and levels of noise and the network configurations we used. The complete results obtained for all data sets and the comparison with the state-of-the-art models are shown in Section~\ref{sec:results} and Section~\ref{sec:comparison}, respectively. We also compared our algorithm with an algorithm that supposes the noise rate is known and we show the results in Section~\ref{sec:compSELFIE}. Then, in Section~\ref{sec:effectiveness}, we analyse the effectiveness of the RAFNI algorithm on two of the data sets used in this study. Finally, we give some final conclusions in Section~\ref{sec:conclusions}.

\section{Background}
\label{sec:background}
In this section, we provide some background in the context of label noise and the types of noise we use in this study (Subsection~\ref{subsec:labelNoiseDef}). Then, we present an overview of the most popular approaches made in the context of deep learning to overcome the problem of label noise and provide a description of the proposals we have selected to compare with RAFNI (Subsection~\ref{subsec:labelNoiseDL}).

\subsection{Definition and types of label noise}
\label{subsec:labelNoiseDef}
In the context of supervised classification, we have a set $\mathbf{X} = \{\mathbf{x}_1, \dots, \mathbf{x}_n\}$ of $n$ training instances and their corresponding labels $\mathbf{y} = (y_1, \dots, y_n)$, where $y_i \in \{1, 2, \dots, K\}$, $i = 1, \dots, n$ and $K$ is the total number of classes. Label noise presents when some instances in $\mathbf{X}$ have erroneous labels. That is, an instance $\mathbf{x}_i \in \mathbf{X}$ with a true label $y_i$ actually appears in the training set with another label $\widetilde{y}_i$, $y_i \neq \widetilde{y}_i$, $i \in \{1, \dots, n\}$. The percentage of instances that present label noise is called the noise rate or noise level.

Depending on whether the label noise appears dependent or independent of the class of the instances and the instances themselves, we can distinguish between the following types of noise:
\begin{itemize}
	\item Symmetric noise (also called uniform noise or NCAR). The noise is independent of the original true class of the instances and the attributes of the instances. Thus, the labels of a percentage of the instances of the training set are randomly changed to another class following a uniform distribution, where all the classes have the same probability of being the noisy label. This implies that the percentage of noisy instances is the same in all classes. There are two options for this type of noise: the noisy label is chosen from the set of all of the classes (thus existing the possibility of not changing the label), or the noisy label is chosen from the set of all classes except the original one. We chose the second option.
	\item Asymmetric noise (also called NAR). The noise is dependent on the original true class of the instances and independent of the attributes of the instances. Therefore, the probability of each class to be the noisy label is different and depends on the original true class, but all the instances in the same class have the same probability of being noisy. This implies that the percentage of noisy instances in each class can be different.
	\item NNAR. The noise is dependent on the original true class of the instances and on the attributes of the instances. Hence, the probability of each class to be the noisy label can be different, depends on the original class, and the instances in each class have different probabilities to be noisy as it depends on the instances themselves.
\end{itemize}


As it happens in classical machine learning, the types of noise that we used can be treated (either by filtering or relabelling) without a prior estimation of the probability distribution.

\subsection{Label noise with deep learning}
\label{subsec:labelNoiseDL}
During the last few years, there has been an increment in the number of proposals to help deep neural networks, and CNNs in particular, to learn in the presence of label noise in supervised classification. Most works fall into one or more of the following approaches:
\begin{itemize}
	\item Proposals that modify the loss function in some way, either to make the loss function robust to label noise \cite{NEURIPS2019_8cd7775f, ghosh2017robust, zhang2018generalized}, or to correct its values, so the noisy labels do not negatively impact the learning \cite{patrini2017making, yi2019probabilistic, ma2018dimensionality, song2019selfie, arazo2019unsupervised}.
	\item Proposals that create a specific deep network architecture \cite{xiao2015learning} or modify an existing one by adding a noise adaptation layer at the end of the desired architecture to model the noise \cite{c51d68a3106242f08ed001d0c46320b3, jindal2016learning}.
	\item Proposals that try to correct the noisy instances \cite{patrini2017making, song2019selfie}.
\end{itemize}

Some proposals suppose that a subset of clean samples is available \cite{xiao2015learning}, and others assume that the noise rate is known \cite{patrini2017making, song2019selfie}, which is not usual when dealing with real-world noisy datasets, though in \cite{patrini2017making} the authors propose a mechanism to approximate the noise rate. A more in-depth survey of all the work that has been done to learn deep neural networks in presence of label noise can be found in \cite{song2020learning}. However, it is important to note that the majority of these proposals are highly focused on classifying the available benchmarks (such as CIFAR10/100 or TinyImagenet), and they use specific networks designed for CIFAR (ResNet32 or ResNet44) along with specific learning schedules. As a result, they are sometimes not generalizable to real-world problems.

We have selected a subset of five of these proposals to compare with our algorithm: one that uses a robust loss function, three that propose loss correction approaches, and one that proposes a hybrid approach between loss correction and sample selection. We choose them because they have official public implementations either on TensorFlow/Keras or PyTorch. In the following, we describe these five proposals.
\begin{enumerate}
	\item Robust loss function approach that uses a generalization of the softmax layer and the categorical cross-entropy loss \cite{NEURIPS2019_8cd7775f}. Here, the authors propose to make the loss function robust against label noise by modifying the loss function and the last softmax activation of the deep neural network with two temperatures, creating non-convex loss functions. These two temperatures can be tuned for each data set. This proposal has the advantage that using the code provided by the authors, it can easily be used with any combination of a deep network, data set and optimization technique, including transfer learning.
	\item Loss correction approaches \cite{patrini2017making}. The authors propose two approaches to correct the loss values of the noisy instances, for which it is necessary to know the noise matrix of the data set, called backward correction and forward correction. They provide a mechanism to estimate the noise matrix, and when used, the approaches are called estimated backward correction and estimated forward correction. The first one uses the noise matrix to correct the loss values, so they are more similar to the loss values of the clean instances. The second explicitly uses the noise matrix to correct the predictions of the model.
	\item Loss correction approach using the dimensionality of the training set \cite{ma2018dimensionality}. The authors explain that, when dealing with noisy labels, the learning can be separated into two phases. In the first phase, which occurs in the first epochs of the training, the network learns the underlying distribution of the data. Then, in the second phase, the network learns to overfit the noisy instances. They use a measure called Local Intrinsic Dimensionality (LID) to detect the moment the training enters the second phase and use the LID to modify the loss function to reduce the effect of the noisy instances.
	\item Loss correction approach \cite{arazo2019unsupervised}, which is based on the static hard bootstrapping loss proposed in \cite{reed2014training} combined with a data augmentation technique called mixup proposed in \cite{zhang2017mixup}. They use a beta mixture model to fit the loss values of the instances so they can distinguish between clean and noisy instances and use the loss correction approach on the noisy ones.
	\item Loss and label correction approach \cite{song2019selfie}. The authors propose a hybrid approach between sample selection and loss correction that tries to relabel noisy instances when possible and not use them when not. For the noisy instances, they rely on the network: if it returns the same label with a high probability in the first epochs of the training, it is possible to correct that instance and the algorithm changes its label to the one the network predicts. In contrast, if the network changes the prediction of an instance inconsistently, they stop using that instance. They assume that the noise rate in the data set is known, and they do not provide a way to estimate it. This approach can be used iteratively so that the training set is iteratively cleaned in several training processes.
\end{enumerate}

\section{RAFNI: Relabelling and filtering instances based on the predictions of the backbone network}
\label{sec:proposal}
In this section, we describe our proposal. First, in Subsection~\ref{subsec:baseConcepts}, we give an overall description of the algorithm and explain its basics. Then, in Subsection~\ref{subsec:formalDef}, we present a formal definition of the algorithm. Finally, in Subsection~\ref{subsec:hyperparameterGuide} we give a guide on how to tune the hyperparameters of the algorithm.

\subsection{Base concepts}
\label{subsec:baseConcepts}

\begin{figure*}[!t]
\centering
\subfloat[\small The standard training process of the backbone network for an epoch $m$]{
  \includegraphics[width=0.8\textwidth, trim=3.2cm 23.5cm 1cm 2.5cm, clip]{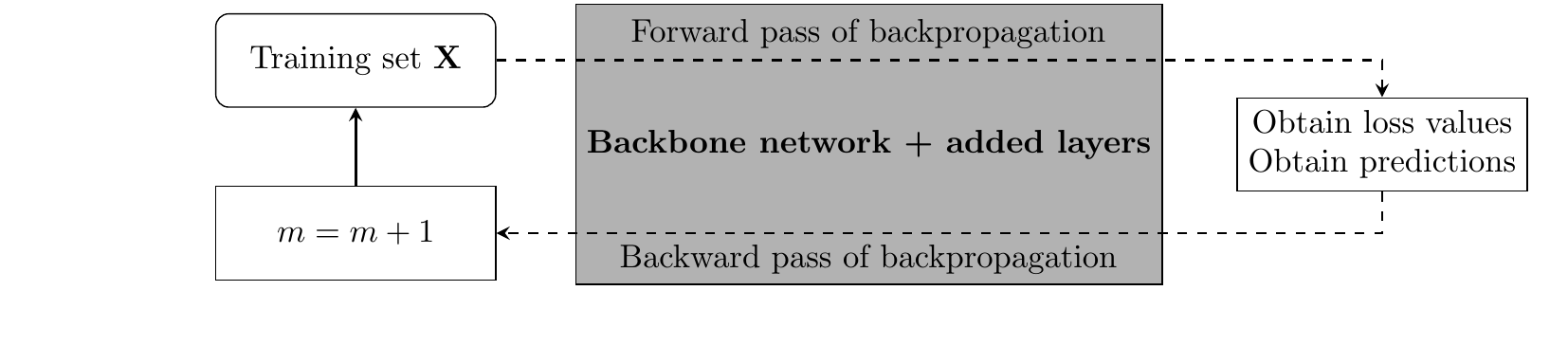}
}

\subfloat[\small The training process of the backbone network using RAFNI for an epoch $m$]{
  \includegraphics[width=0.8\textwidth, trim=3.2cm 19.3cm 1cm 6.7cm, clip]{training}
}
\caption{Difference between training the backbone network (a) without and (b) with the RAFNI algorithm}
\label{fig:training}
\end{figure*}

We propose the RAFNI algorithm, which filters and relabels instances based on the predictions and their probabilities made by the backbone neural network during the training process. In Figure~\ref{fig:training}, we show the difference between training the backbone network with and without the RAFNI algorithm and the moment it is applied. The backbone network used is independent of the algorithm, and it can change or be modified, for example, including transfer learning.

Generally speaking, we propose two mechanisms to filter an instance and one mechanism to relabel an instance, with some restrictions. These mechanisms are the following:
\begin{itemize}
	\item First filtering mechanism. This mechanism only uses the loss value of the instances. The foundation is that the noisy instances tend to have higher loss values than the rest of them. As a result, this mechanism filters out instances that have a loss value above a certain threshold. This threshold is dynamic and will change during training.
	\item Second filtering mechanism. This mechanism depends on how many times the backbone network has changed the prediction of an instance. Here we suppose that if the backbone network changes its prediction too many times is because it is unsure about its class and it is better to remove that instance. Thus, this mechanism filters an instance if its predictions change more than a certain number of times.
	\item Relabelling mechanism. This mechanism takes into account the probability predictions of the backbone network. We suppose that if the backbone network predicts another class with a high probability as the training progresses, it is probable that the instance is noisy and its class is indeed the one predicted by the backbone network. As a consequence, the relabelling mechanism changes the class of an instance if the backbone network predicts another class with a probability that is above a certain threshold. This threshold is also dynamic and will change during training. In addition, we establish a period of a certain number of epochs after an instance has been relabelled during which the algorithm cannot filter nor relabel it again.
\end{itemize}

These mechanisms have restrictions related to the moment they are applied. Since we are using the backbone network to relabel and filter instances, we need to wait until the network is sufficiently trained for the predictions to be reliable. This can be measured using the loss values of the instances. Intuitively, we want to start the algorithm (and thus the three mechanisms) when the backbone network has learned to classify the clean instances but it has not learned to overfit the noisy ones yet. Here, similarly to \cite{arazo2019unsupervised}, we approximated the loss values of the instances in each epoch of the training process by a mixture model with two components, but in our case, we use a Gaussian mixture model. To do that, we used the expectation minimization algorithm and used the two components to detect the moment where the RAFNI algorithm needs to start.

A mixture model is a model that can represent different subpopulations inside a population. These subpopulations or components follow a distribution that in a Gaussian mixture model is supposed to be a Gaussian distribution. That way, if we have a Gaussian mixture model with two components, we are approximating two subpopulations, each one with a Gaussian distribution, so we obtain two means and two variances. In our case, we have two components, one for the clean instances, with mean $\mu_{\mathrm{clean}}$ and standard deviation $\sigma_{\mathrm{clean}}$, and one for the noisy instances, with mean $\mu_{\mathrm{noisy}}$ and standard deviation $\sigma_{\mathrm{noisy}}$.

We detect the moment we need to start the RAFNI algorithm by calculating the overlap between the two Gaussians obtained by the mixture model over the loss values of the instances in all training epochs. At first, the two Gaussians will start separating from one another, while the network learns to classify the clean and easy examples. Then, at some point, they will start to get closer, as the network starts to overfit the noisy examples. Therefore, we start the algorithm when the overlap between the Gaussians is below a fixed value or when this overlap starts to increase. We have tested different values for this hyperparameter with different data sets and levels of noise and we found that 0.15 is a good value that can remain fixed across all data sets and noise rates.

\subsection{Formal definition}
\label{subsec:formalDef}

Let $\mathbf{X} = \{\mathbf{x}_1, \mathbf{x}_2, \dots, \mathbf{x}_n\}$ be the set of $n$ training instances and $\mathbf{y} = (y_1, y_2, \dots,$ $ y_n)$ their corresponding labels, where $y_i \in \{1, 2, \dots, K\}$, $i = 1, \dots, n$, and $K$ is the total number of classes. Let $m$ be an epoch of the training process, $m = 1, \dots, M$, where $M$ is the total number of epochs, and $\mathbf{l}_m = (l_{m1}, l_{m2}, \dots, l_{mn})$ the losses of the training instances in epoch $m$, $m = 1, \dots, M$. Finally, let $\mathbf{p}_{mi} = (p_{mi1}, p_{mi2}, \dots, p_{miK})$ be the probabilities predicted by the backbone neural network for each instance $\mathbf{x}_i \in \mathbf{X}$ in epoch $m$, and $\widehat{y}_{mi} \in \{1, 2, \dots, K\}$ the prediction of the backbone network for the instance $\mathbf{x}_i$ in epoch $m$, where $m = 1, \dots, M$, $i = 1, \dots, n$. Then, we define the following:
\begin{enumerate}
	\item A threshold, named \texttt{loss\_threshold}, so that if $l_{mi} >$ \texttt{loss\_threshold}, then $\mathbf{x_i}$ is removed from the training set for the following epochs $m+1, \dots, M$, where $i =1, \dots, n$.
	\item A number, \texttt{record\_length}, denoting the length of the \texttt{record} of each instance, so that the algorithm saves the last \texttt{record\_length} predictions made by the neural network in the last \texttt{record\_length} epochs of the training. Then, if the predictions of an instance $\mathbf{x}_i$ change \texttt{record\_length}$-1$ times in the last \texttt{record\_length} epochs, the instance $\mathbf{x}_i$ is removed from the training set for the following epochs $m+1, \dots, M$, where $i = 1, \dots, n$.
	\item A threshold, \texttt{prob\_threshold}, so that if $\max_k (\mathbf{p}_{mi}) >$ \texttt{prob\_threshold} and $y_i \neq \widehat{y}_{mi}$, then $y_i = \widehat{y}_{mi}$ in the following epochs $m+1, \dots, M$, where $i = 1, \dots, n$. If this happens, the algorithm clears the \texttt{record} of the instance $\mathbf{x}_i$.
	\item A number, \texttt{not\_change\_epochs}, so that if the label of an instance has been changed, the algorithm cannot change it again nor remove that instance from the training set until \texttt{not\_change\_epochs} epochs have passed.
\end{enumerate}

\begin{figure*}[!t]
\begin{adjustbox}{center}
\includegraphics[width=0.8\textwidth, trim= 3.2cm 6.5cm 1.9cm 2.5cm, clip]{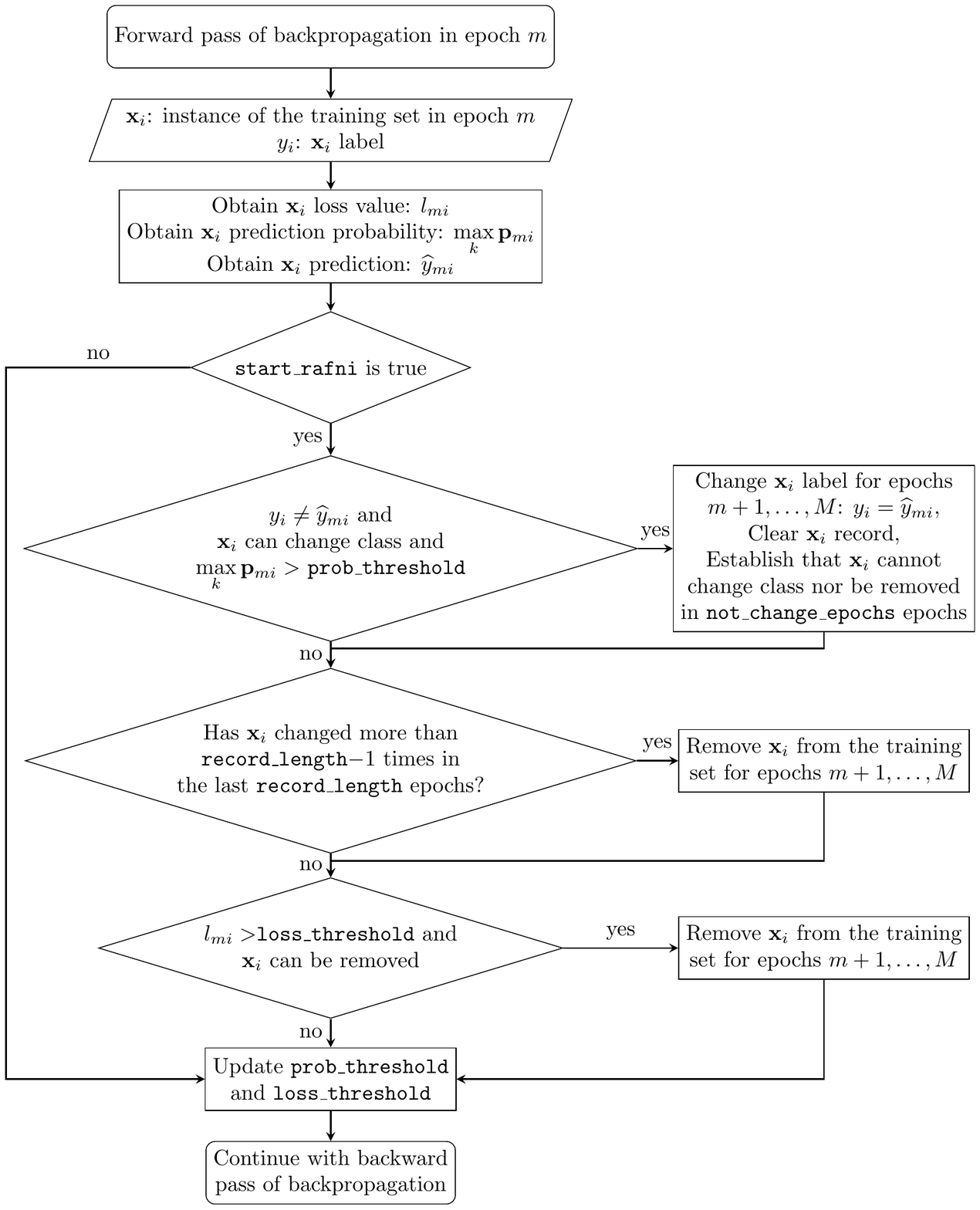}
\end{adjustbox}
\caption{Flowchart of the RAFNI algorithm}
\label{fig:flowchart}
\end{figure*}

In Figure~\ref{fig:flowchart}, we show the flowchart of the RAFNI algorithm, detailing how and when each mechanism is applied to each instance $\mathbf{x}_i$ during a specific epoch $m$ of the training process.

The numbers \texttt{record\_length} and \texttt{not\_change\_epochs} are hyperparameters of the algorithm that can be setted by the user. The two thresholds, \texttt{loss\_threshold} and \texttt{prob\_threshold}, are parameters that dynamically change every epoch $m$ using the losses of the instances and their probabilities in the previous epoch, $l_{m-1}$. Specifically, the \texttt{loss\_threshold} is calculated for every epoch as the quantile of order $x_1$ ($x_1$ is an hyperparameter of the algorithm called \texttt{quantile\_loss}, that can be setted by the user) of the losses in the previous epoch $l_{m-1}$. Similarly, the \texttt{prob\_threshold} is calculated for every epoch as the quantile of order $x_2$ (also a hyperparameter, called \texttt{quantile\_prob}) of the probabilities returned by the backbone network for the misclassified instances. That way, the \texttt{loss\_threshold} usually descends as the epoch increases and the training instances are being filtered and their classes relabelled. Due to that, we need to stop updating the \texttt{loss\_threshold} parameter at some point to not filter too many instances. Similarly, we also need to stop updating the \texttt{prob\_threshold} parameter so it does not change the class of too many instances. To do this, we use again the two Gaussians obtained by the Gaussian mixture model and stop the update of both thresholds when the means of the Gaussians are sufficiently close. We tested different values and we obtained that 0.3 is a good value that works for different data sets and levels of noise. That way, we stop updating the thresholds if $\mu_{\mathrm{noisy}} - \mu_{\mathrm{clean}} < 0.3$.

This algorithm can be used with any CNN as a backbone network. The code of the algorithm is available at \url{https://github.com/ari-dasci/S-RAFNI}.

\subsection{A guide to the hyperparameters of RAFNI}
\label{subsec:hyperparameterGuide}

RAFNI has a list of hyperparameters that can be fine-tuned by the user. Here we specify which hyperparameters are most important to be tuned if a validation set is available, which ones are less important and which ones do not need to be tuned. We also give a guide on how to tune them.

The complete list of hyperparameters of RAFNI is the following: the overlapping threshold between the noisy Gaussian and the clean Gaussian we use to start the algorithm, the difference between the means of the two Gaussians we use to stop the update of the \texttt{loss\_threshold} and the \texttt{prob\_threshold}, the \texttt{quantile\_loss}, the \texttt{quantile\_prob}, the \texttt{record\_length} and the \texttt{not\_change\_epochs}. The \texttt{loss\_threshold} and the \texttt{prob\_threshold} are not really hyperparemeters as they cannot be tuned, they change dinamically in each epoch based on the \texttt{quantile\_loss} and \texttt{quantile\_prob} hyperparameters, respectively.

We also have the hyperparameters inherent to training a deep neural network: the total number of epochs of the training, the batch size and whether to use fine-tune or not in the backbone CNN: if we do not use fine-tuning, the layers of the backbone neural network are not retrained and only the new added layers are trained, and if we use fine-tuning, all the layers are trained. Whether to use fine-tuning or not depends on the backbone network used (how deep it is) and if the data set we want to classify has enough images to retrain the whole network or not. The number of epochs of the training depends on if we are fine-tuning the backbone network and the size of the data set. The batch size depends on the size of the data set, it will increase as the size of the data set increases, usually.

\begin{figure*}
\centering
\subfloat[Epoch 0]{\includegraphics[width=0.34\textwidth]{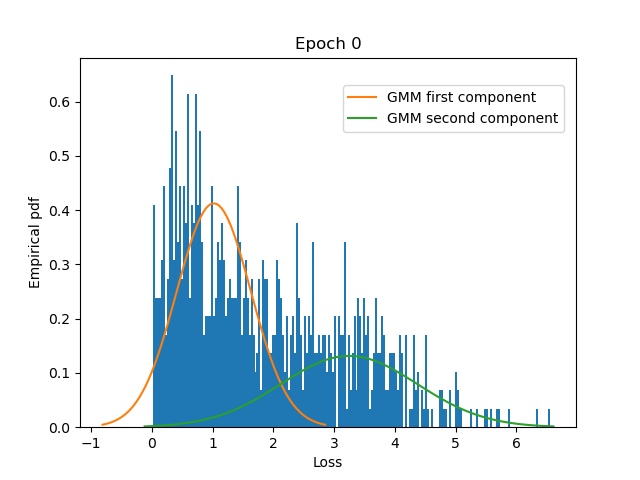}}
\subfloat[Epoch 1]{\includegraphics[width=0.34\textwidth]{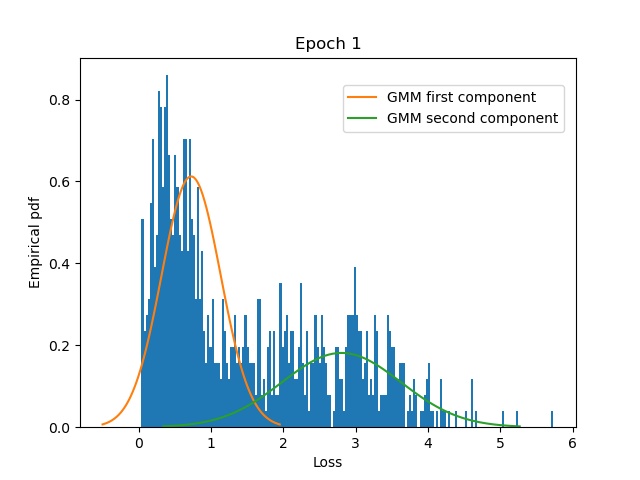}}
\subfloat[Epoch 2]{\includegraphics[width=0.34\textwidth]{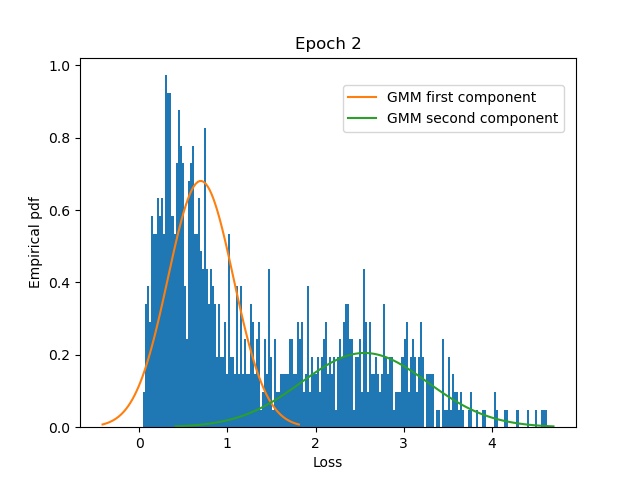}}
\caption{The two components obtained by the Gaussian Mixture Model (GMM) over the loss values of the instances in the first three epochs of the training using the EILAT data set at 40\% of noise.}
\label{fig:GMM}
\end{figure*}

\begin{figure*}
\centering
\subfloat[EILAT at 40\% noise]{\includegraphics[width=0.34\textwidth]{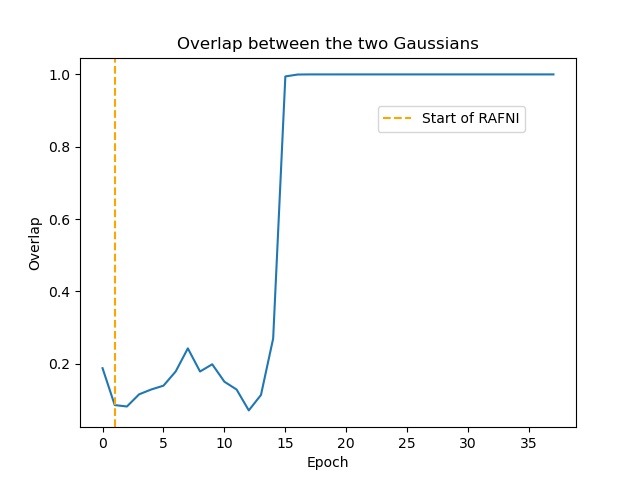}}
\subfloat[StructureRSMAS at 10\% noise]{\includegraphics[width=0.34\textwidth]{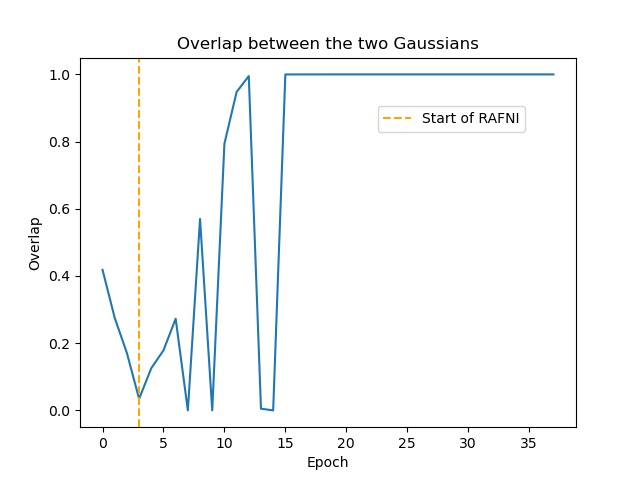}}
\subfloat[RSMAS at 30\% noise]{\includegraphics[width=0.34\textwidth]{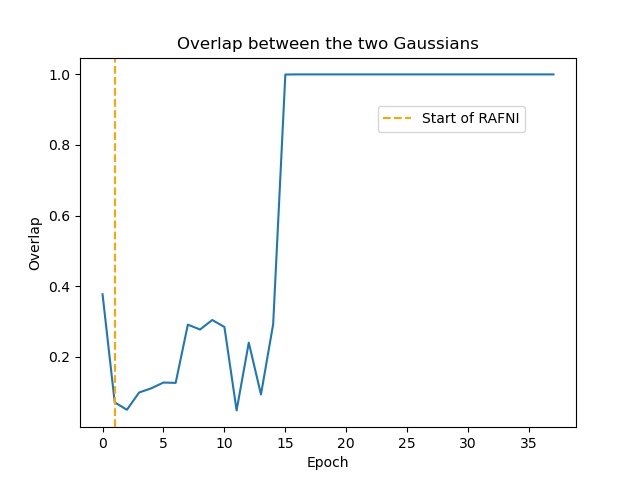}}
\caption{Evolution of the overlap between the two components of the GMM through the epochs of the training of different data sets and noise rates.}
\label{fig:overlap}
\end{figure*}

\begin{figure*}
\centering
\subfloat[EILAT at 70\% noise]{\includegraphics[width=0.34\textwidth]{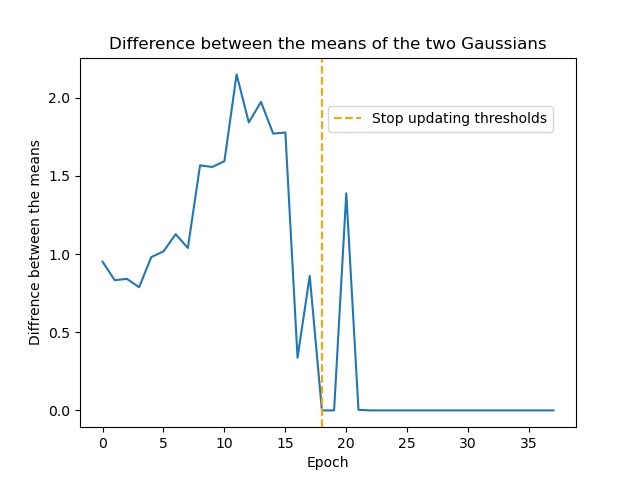}}
\subfloat[StructureRSMAS at 10\% noise]{\includegraphics[width=0.34\textwidth]{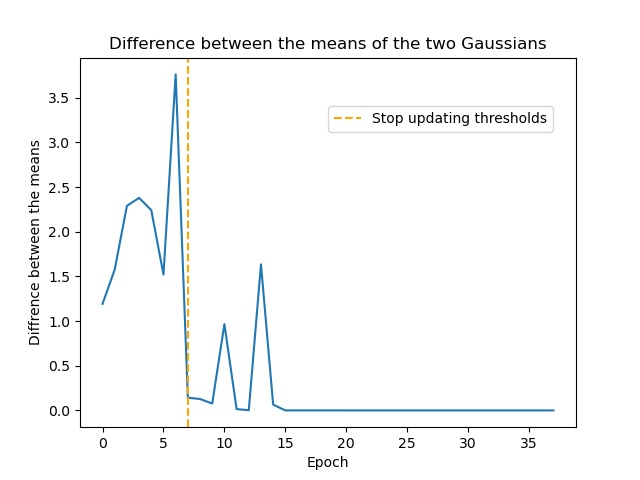}}
\subfloat[RSMAS at 30\% noise]{\includegraphics[width=0.34\textwidth]{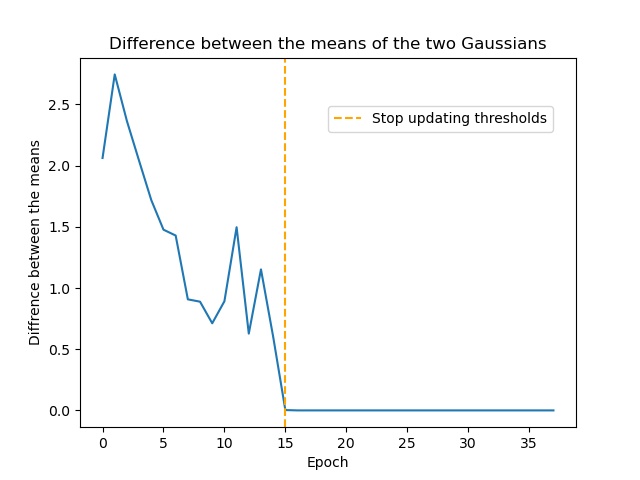}}
\caption{Evolution of the difference between the means of the two components of the GMM through the epochs of the training of different data sets and noise rates.}
\label{fig:difference}
\end{figure*}

If we focus on the specific hyperparameters of RAFNI, there are two of them that we recommend not changing: the overlapping threshold between the Gaussians we use to start the algorithm and the difference between the means of the Gaussians that we use to stop the updates of \texttt{loss\_threshold} and \texttt{prob\_threshold}. We tested different values for these parameters across all the data sets and levels of noise we used and we found that 0.15 is a good value for the overlapping threshold and 0.3 a good value for the difference between the means of the Gaussians. To show why we chose these values we can see Figures~\ref{fig:GMM}, \ref{fig:overlap} and \ref{fig:difference}. In Figure~\ref{fig:GMM} we show the two components (the two Gaussians) obtained by the Gaussian Mixture Model (GMM) over the losses of the instances in the first epochs of the training of EILAT with 40\% of symmetric noise. At first, as the learning progresses, the network starts to differentiate between the clean and the noisy instances, and thus the two components start to separate from themselves. Then, the network starts to overfit the noisy instances and the two components start to come together again. To stop this overfitting, we start the RAFNI algorithm when the overlap between the two components is less than 0.15 or when the overlap in an epoch is greater than in the previous epoch. In Figure~\ref{fig:overlap} we can see how this overlap changes through the epochs of the training and in which specific epoch we are starting the RAFNI algorithm. Finally, to stop the updating of the \texttt{loss\_threshold} and the \texttt{prob\_threshold} we use the distance between the means of the two components: if they are close, it means that there are not enough noisy instances, so the two components are very close. In Figure~\ref{fig:difference} we can see the evolution of the difference between the means of the two Gaussians and the specific epoch where we stop the updating of the thresholds.

Regarding the rest of the hyperparameters of RAFNI, we recommend to tune them if possible, though we found that tunning the \texttt{quantile\_loss} and \texttt{quantile\_prob} is more important than tunning the \texttt{record\_length} and \texttt{not\_change\_epochs} hyperparameters. The \texttt{quantile\_loss} depends on how hard is the data set to classify: the more difficult it is, the less we can rely on the predictions of the backbone network and it is more convenient to use higher values so that the algorithm is more conservative, that is, so it does not remove nor change the class of too many instances. Something similar happens with the values of the hyperparameters \texttt{quantile\_prob}, \texttt{record\_length} and \texttt{not\_change\_epochs}, though in these cases is also related to the size of the data set: the harder it is to classify and fewer images have, higher values we should give to these hyperparameters. In the case of \texttt{record\_length} and \texttt{not\_change\_epochs}, we should also take into account the total number of epochs of the training: if this number is small, these two hyperparameters should also be small and they should not exceed the total number of epochs in any case).

The ranges in which each hyperparameter can take values are the following. For the \texttt{quantile\_loss} and the \texttt{quantile\_prob} hyperparameters, given they are quantiles, their maximum value is 1 and their minimum is 0. However, we found that they perform best if they vary in the range $[0.6, 0.99]$. The minimum value for \texttt{record\_length} is 2, so it can track at least one change in the class of the instances, and its maximum is the total number of epochs in the training; the higher this value is, the fewer instances the algorithm will remove because their class has changed. Finally, the minimum value for \texttt{not\_change\_epochs} is 1 and the maximum is the total number of epochs in the training.

\section{Experimental framework}
\label{sec:experimentalFramework}

In this section, we describe the experimental framework we used to carry out the experiments. In Subsection~\ref{subsec:datasets}, we describe the data sets we used. In Subsection~\ref{subsec:noise}, we detail the types of noise we used in each data set along with the noise levels we used in each one of them. Finally, in Subsection~\ref{subsec:configuration}, we provide the specific configuration, backbone neural network and software we used for all the experiments.

\subsection{Data sets}
\label{subsec:datasets}

\begin{table}[]
\centering
\caption{A summary of the data sets used in this study.}
\label{tab:datasets}
\begin{tabular}{|c|c|c|c|}
\hline
\textbf{Data set}       & \textbf{\# classes} & \textbf{\# images} & \textbf{\# images per class}                                                                                                                                  \\ \hline
RSMAS          & 14         & 766       & \begin{tabular}[c]{@{}c@{}}1: 109, 2: 77, 3: 57, 4: 63,\\ 5: 24, 6: 60, 7: 22, 8: 79,\\ 9: 54, 10: 28, 11: 32,\\ 12: 88, 13: 37, 14: 36\end{tabular} \\ \hline
StructureRSMAS & 14         & 409       & \begin{tabular}[c]{@{}c@{}}1: 44, 2: 41, 3: 34, 4: 20,\\ 5: 16, 6: 18, 7: 33, 8: 38,\\ 9: 30, 10: 21, 11: 32,\\ 12: 23, 13: 36, 14: 23\end{tabular}  \\ \hline
EILAT          & 8          & 1123      & \begin{tabular}[c]{@{}c@{}}1: 87, 2: 78, 3: 29, \\ 4: 160, 5: 200, 6: 216, \\ 7: 296, 8: 11.\end{tabular}                                            \\ \hline
COVIDGR1.0-SN  & 2          & 700       & 350                                                                                                                                                  \\ \hline
CIFAR10        & 10         & 60000     & 6000                                                                                                                                                 \\ \hline
CIFAR100       & 100        & 60000     & 600                                                                                                                                                  \\ \hline
\end{tabular}
\end{table}

We describe the data sets we used to analyse RAFNI under different types and levels of label noise. We used six data sets, each one with a different number of classes, images per class and a total number of images: RSMAS, StructureRSMAS, EILAT, COVIDGR1.0-SN, CIFAR10 and CIFAR100. There is a summary of the statistics of these data sets in Table~\ref{tab:datasets}.

RSMAS, StructureRSMAS and EILAT are small coral data sets. RSMAS and EILAT \cite{coralDataset, GOMEZRIOS2019315} are texture data sets, containing coral patches, meaning that they are close-up patches extracted from larger images, and StructureRSMAS \cite{GOMEZRIOS2019104891} is a structure data set, containing images of entire corals. The patches in EILAT have a size of 64$\times$64 and come from images taken under similar underwater conditions, and the ones in RSMAS have a size of 256$\times$256 and come from images taken under different conditions. StructureRSMAS is a data set collected from the Internet and therefore contains images of different sizes taken under different conditions.

COVIDGR1.0-SN is a modification of COVIDGR1.0 \cite{9254002}. COVIDGR1.0 contains chest x-rays of patients divided into two classes: positive for COVID-19, and negative for COVID-19, using the RT-PCR as ground truth. All the images in the data set were taken using the same protocol and similar x-ray machines. The authors made available the data set along with a list containing the degree of severity of the positive x-rays: Normal-PCR+, Mild, Moderate and Severe. The x-rays with Normal-PCR+ severity are x-rays from patients that tested positive in the RT-PCR test but where expert radiologists could not find signs of the disease in the x-ray. The modification that we use, COVIDGR1.0-SN, is the same data set as COVIDGR1.0 but we removed the 76 positive images with Normal-PCR+ severity. To maintain the two classes balanced, as happens in the original data set, we also removed 76 randomly chosen negative images.

Finally, CIFAR10 and CIFAR100 \cite{krizhevsky2009learning} are the 60k tiny images of size 32$\times$32 images proposed by Alex Krizhevsky. Concerning the other data sets used in this study, CIFAR10 and CIFAR100 are much larger in size. Both of them have a predefined test hold-out of 10.000 images, meaning they both have a training set of 50.000 images. Both datasets contain classes of common objects, such as 'Airplane' and 'Ship' in CIFAR10 or 'Bed' and 'Lion' in CIFAR100.

\subsection{Types and levels of label noise}
\label{subsec:noise}

\begin{table}[]
\centering
\caption{Types and levels of label noise used for each data set}
\label{tab:noise}
\begin{tabular}{|c|c|c|}
\hline
\textbf{Data set}       & \textbf{Type of label noise}              & \textbf{Levels of noise}                                                                                       \\ \hline
RSMAS          & \multirow{3}{*}{Symmetric} & \multirow{3}{*}{\begin{tabular}[c]{@{}c@{}}0\%, 20\%, 30\%, 40\%,\\ 50\%, 60\% and 70\%\end{tabular}} \\ \cline{1-1}
StructureRSMAS &                                  &                                                                                                       \\ \cline{1-1}
EILAT          &                                  &                                                                                                       \\ \hline
COVIDGR1.0-SN  & NNAR          & 0\%, 20\%, 30\%, 40\% and 50\%                                                                        \\ \hline
CIFAR10        & Asymmetric                 & 0\%, 20\%, 30\% and 40\%                                                                              \\ \hline
CIFAR10        & \multirow{2}{*}{Symmetric} & \multirow{2}{*}{0\%, 20\%, 40\% and 60\%}                                                             \\ \cline{1-1}
CIFAR100       &                                  &                                                                                                       \\ \hline
\end{tabular}
\end{table}

We state the types of noise we used for each data set and which rates we used for each one of them. In Table~\ref{tab:noise} we show a brief description. In summary, we used symmetric noise, asymmetric noise and NNAR noise. Symmetric noise is the most used type of noise and since it is not necessary to have external information to use it, we use this type of noise in all data sets except for COVIDGR1.0-SN. But, to also use more realistic and challenging types of noise, we used asymmetric and NNAR noise when possible. This is the case for CIFAR10 and COVIDGR1.0-SN.

For CIFAR10, we used the asymmetric noise introduced in \cite{patrini2017making}, which has been a standard when evaluating deep learning in the presence of asymmetric label noise. This noise is introduced between classes that are alike, simulating real label noise that could have occurred naturally. In particular, we introduced asymmetric noise between the following classes: TRUCK $\rightarrow$ AUTOMOBILE, BIRD $\rightarrow$ AIRPLANE, DEER $\rightarrow$ HORSE, CAT $\leftrightarrow$ DOG, as defined in \cite{patrini2017making}. Note that since we are introducing an $x$\% of noise in five of the ten classes, we are introducing an $\frac{x}{2}$\% of noise in the total dataset.

For COVIDGR1.0-SN, we have the additional information on the severity degree in the images of the positive class, so we used them to introduce NNAR noise, where we change the labels of a percentage $x$ of the instances of the data set subject to some condition over the instances. COVIDGR1.0-SN has two classes: P (COVID-19 positive) and N (COVID-19 negative), and the instances from P have associated a severity (Mild, Moderate and Severe). In this scenario, is more realistic that a positive image with mild severity has been misclassified as negative than a positive image with moderate or severe severity. Equivalently, is more realistic that a positive image with moderate severity has been misclassified as negative than a positive image with severe severity. As a consequence, we define the probability of the instances in the groups N (to change class to P), Mild (to change class to N), Moderate (to change class to N) and Severe (to change class to N) as it follows: 0.5 for N, 0.3 for Mild, 0.2 for Moderate and 0 for Severe. That way, we are changing the same amount of instances from P to N and vice versa, but when we change the class from P to N, it is more probable to change a mild positive image than a moderate positive image. In addition, we are making sure that no positive image with severe severity has changed from class P to N.

\subsection{Network and experimental configuration}
\label{subsec:configuration}

Here, we provide the specific configuration we used in the experiments we carried out.

\begin{table}[]
\centering
\caption{Fixed hyperparameters we used in each data set.}
\label{tab:fixedHyper}
\begin{tabular}{|c|c|c|c|c|}
\hline
Data set       & Optimizer & Batch size & \begin{tabular}[c]{@{}c@{}}Total number\\ of epochs\end{tabular} & Fine-tune \\ \hline
RSMAS          & SGD       & 16         & 40                                                               & No        \\ \hline
StructureRSMAS & SGD       & 16         & 40                                                               & No        \\ \hline
EILAT          & SGD       & 16         & 40                                                               & No        \\ \hline
COVIDGR1.0-SN     & SGD       & 16         & 40                                                               & No        \\ \hline
CIFAR10        & SGD       & 128         & 10                                                              & Yes       \\ \hline
CIFAR100       & SGD       & 128         & 15                                                              & Yes       \\ \hline
\end{tabular}
\end{table}

As the backbone neural network we used ResNet50 \cite{he2016deep}, though the implementation of RAFNI is independent of the backbone network and can be changed easily. We used ResNet50 pre-trained using ImageNet, removing the last layer of the network and adding two fully connected layers, the first one with 512 neurons and ReLU activation and the second one with as many neurons as classes the data set had and softmax activation. Once we removed the last layer of ResNet50, its output had 2048 neurons. We chose 512 neurons for the first fully connected layer we added as an intermediate number between 2048 and the number of classes in the data sets. The fixed hyperparameters we used in each data set can be seen in Table~\ref{tab:fixedHyper}. We used the Stochastic Gradient Descent (SGD) with a learning rate of $1\times10^{-3}$, a decay of $1\times10^{-6}$ and a Nesterov momentum of 0.9. We did not optimize these hyperparameters.

\begin{table}[]
\centering
\caption{Values we used for each hyperparameter and data set for the grid search.}
\label{tab:valuesHyper}
\begin{tabular}{|c|c|c|}
\hline
Data set                                                                                                    & Hyperparameter      & Values                                \\ \hline
\multirow{4}{*}{\begin{tabular}[c]{@{}c@{}}RSMAS, EILAT,\\ StructureRSMAS\\ and\\ COVIDGR1.0-SN\end{tabular}} & \texttt{quantile\_loss}      & \{0.9, 0.92, 0.94, 0.96, 0.98, 0.99\} \\ \cline{2-3} 
                                                                                                            & \texttt{quantile\_prob}      & \{0.9, 0.93, 0.95, 0.97, 0.99\}       \\ \cline{2-3} 
                                                                                                            & \texttt{record\_length}      & \{5, 8\}                              \\ \cline{2-3} 
                                                                                                            & \texttt{not\_change\_epochs} & \{4, 7\}                              \\ \hline
\multirow{4}{*}{\begin{tabular}[c]{@{}c@{}}CIFAR10\end{tabular}}                            & \texttt{quantile\_loss}      & \{0.95, 0.97, 0.99\}            \\ \cline{2-3} 
                                                                                                            & \texttt{quantile\_prob}      & \{0.75, 0.8, 0.85, 0.9\}       \\ \cline{2-3} 
                                                                                                            & \texttt{record\_length}      & \{2, 4\}                              \\ \cline{2-3} 
                                                                                                            & \texttt{not\_change\_epochs} & \{1, 2\}                              \\ \hline
                                                                                                            \multirow{4}{*}{\begin{tabular}[c]{@{}c@{}}CIFAR100\end{tabular}}                            & \texttt{quantile\_loss}      & \{0.95, 0.97, 0.99\}            \\ \cline{2-3} 
                                                                                                            & \texttt{quantile\_prob}      & \{0.75, 0.78, 0.82, 0.85\}       \\ \cline{2-3} 
                                                                                                            & \texttt{record\_length}      & \{2, 4\}                              \\ \cline{2-3} 
                                                                                                            & \texttt{not\_change\_epochs} & \{1, 2\}                              \\ \hline
\end{tabular}
\end{table}

For the experimentation, we used TensorFlow 2.4 and an Nvidia Tesla V100. The values we gave to each hyperparameter can be seen in Table~\ref{tab:valuesHyper}. Using those values, we performed an exhaustive grid search to find the best configuration in each case. Since the CIFAR data sets and the rest of the data sets we used had different sizes, the experimental framework we used for them was different. For the smaller data sets RSMAS, EILAT, StructureRSMAS and COVIDGR1.0-SN, we used five-fold cross-validation for the experiments in the grid search, while for CIFAR10 and CIFAR100 we used a 20\% hold-out using only the original train set. Then, to ensure a more stable final result, we did the following. For the smaller data sets, we repeated the five-fold cross-validation with the best hyperparameter configuration five times (noted 5x5fcv) and we report the mean and standard deviation of the 5x5fcv. This scheme of using mean and standard deviation is one of the most used in the literature. For the CIFAR data sets, we used the best hyperparameter configuration (found using only the training set) in the predefined test hold-out, we repeated that experiment five times and we report the mean and standard deviation.

We used the accuracy measure, widely used for supervised classification. The accuracy is defined as the number of instances well classified in the test set divided by the total number of instances in the test set.

To make the comparison with the baseline model (that is, the backbone network alone, without filtering or relabelling instances), we used the same scheme as we used with RAFNI: we repeated five times the five-fold cross-validation (or the hold-out for the CIFAR data sets) and we report mean and standard deviation.

\begin{table*}[]
\centering
\caption{Best hyperparameter values for all data sets using ResNet50.}
\label{tab:bestHyper}
\begin{tabular}{|c|c|c|c|c|c|}
\hline
Data set                                                                             & Noise rate & quantile\_loss & quantile\_prob & record\_length & not\_change\_epochs \\ \hline
\multirow{8}{*}{RSMAS}                                                               & 0\%        & 0.99           & 0.95           & 5              & 7                   \\ \cline{2-6} 
                                                                                     & 10\%       & 0.96           & 0.93           & 5              & 7                   \\ \cline{2-6} 
                                                                                     & 20\%       & 0.92           & 0.95           & 8              & 4                   \\ \cline{2-6} 
                                                                                     & 30\%       & 0.92           & 0.9            & 5              & 7                   \\ \cline{2-6} 
                                                                                     & 40\%       & 0.9            & 0.95           & 5              & 7                   \\ \cline{2-6} 
                                                                                     & 50\%       & 0.9            & 0.93           & 5              & 7                   \\ \cline{2-6} 
                                                                                     & 60\%       & 0.9            & 0.9            & 8              & 4                   \\ \cline{2-6} 
                                                                                     & 70\%       & 0.9            & 0.93           & 5              & 4                   \\ \hline
\multirow{8}{*}{StructureRSMAS}                                                      & 0\%        & 0.99           & 0.9            & 8              & 4                   \\ \cline{2-6} 
                                                                                     & 10\%       & 0.96           & 0.95           & 8              & 7                   \\ \cline{2-6} 
                                                                                     & 20\%       & 0.94           & 0.95           & 5              & 4                   \\ \cline{2-6} 
                                                                                     & 30\%       & 0.92           & 0.95           & 5              & 4                   \\ \cline{2-6} 
                                                                                     & 40\%       & 0.92           & 0.95           & 5              & 4                   \\ \cline{2-6} 
                                                                                     & 50\%       & 0.92           & 0.95           & 8              & 4                   \\ \cline{2-6} 
                                                                                     & 60\%       & 0.9            & 0.95           & 5              & 4                   \\ \cline{2-6} 
                                                                                     & 70\%       & 0.9            & 0.9            & 5              & 7                   \\ \hline
\multirow{8}{*}{EILAT}                                                               & 0\%        & 0.99           & 0.99           & 8              & 4                   \\ \cline{2-6} 
                                                                                     & 10\%       & 0.98           & 0.97           & 5              & 7                   \\ \cline{2-6} 
                                                                                     & 20\%       & 0.94           & 0.9            & 8              & 7                   \\ \cline{2-6} 
                                                                                     & 30\%       & 0.94           & 0.95           & 8              & 4                   \\ \cline{2-6} 
                                                                                     & 40\%       & 0.92           & 0.97           & 8              & 7                   \\ \cline{2-6} 
                                                                                     & 50\%       & 0.9            & 0.93           & 8              & 4                   \\ \cline{2-6} 
                                                                                     & 60\%       & 0.9            & 0.95           & 5              & 4                   \\ \cline{2-6} 
                                                                                     & 70\%       & 0.9            & 0.9            & 5              & 4                   \\ \hline
\multirow{6}{*}{COVIDGR1.0-SN}                                                       & 0\%        & 0.96           & 0.99           & 8              & 7                   \\ \cline{2-6} 
                                                                                     & 10\%       & 0.99           & 0.9            & 5              & 7                   \\ \cline{2-6} 
                                                                                     & 20\%       & 0.94           & 0.97           & 5              & 4                   \\ \cline{2-6} 
                                                                                     & 30\%       & 0.96           & 0.97           & 5              & 4                   \\ \cline{2-6} 
                                                                                     & 40\%       & 0.92           & 0.93           & 5              & 7                   \\ \cline{2-6} 
                                                                                     & 50\%       & 0.9            & 0.95           & 5              & 7                   \\ \hline
\multirow{4}{*}{\begin{tabular}[c]{@{}c@{}}CIFAR10,\\ symmetric noise\end{tabular}}  & 0\%        & 0.99           & 0.75           & 4              & 1                   \\ \cline{2-6} 
                                                                                     & 20\%       & 0.95           & 0.75           & 2              & 1                   \\ \cline{2-6} 
                                                                                     & 40\%       & 0.97           & 0.75           & 2              & 2                   \\ \cline{2-6} 
                                                                                     & 60\%       & 0.99           & 0.8            & 2              & 2                   \\ \hline
\multirow{4}{*}{\begin{tabular}[c]{@{}c@{}}CIFAR10,\\ asymmetric noise\end{tabular}} & 0\%        & 0.99           & 0.75           & 4              & 1                   \\ \cline{2-6} 
                                                                                     & 20\%       & 0.95           & 0.8            & 2              & 2                   \\ \cline{2-6} 
                                                                                     & 30\%       & 0.95           & 0.85           & 2              & 2                   \\ \cline{2-6} 
                                                                                     & 40\%       & 0.95           & 0.85           & 2              & 1                   \\ \hline
\multirow{4}{*}{\begin{tabular}[c]{@{}c@{}}CIFAR100,\\ symmetric noise\end{tabular}} & 0\%        & 0.95           & 0.85           & 4              & 2                   \\ \cline{2-6} 
                                                                                     & 20\%       & 0.95           & 0.78           & 4              & 1                   \\ \cline{2-6} 
                                                                                     & 40\%       & 0.99           & 0.75           & 2              & 1                   \\ \cline{2-6} 
                                                                                     & 60\%       & 0.99           & 0.75           & 2              & 1                   \\ \hline
\end{tabular}
\end{table*}

The best values for the hyperparameters in each case can be found in Table~\ref{tab:bestHyper}.

\section{Comparison with the baseline model}
\label{sec:results}
In this section, we present the results we obtained for each data set using our proposal, and we compare it with the backbone network alone as the baseline. 

\subsection{RSMAS, EILAT, StructureRSMAS and COVIDGR1.0-SN}
\label{subsec:smallDatasets}

We present the results obtained with the smaller data sets: RSMAS, EILAT, StructureRSMAS and COVIDGR1.0-SN.

\begin{table*}[]
\centering
\caption{5x5fcv mean $\pm$ std accuracy obtained for the data sets RSMAS, EILAT and StructureRSMAS using RAFNI with ResNet50 as backbone network and the backbone network alone, ResNet50, as baseline. The best results in each case are stressed in bold.}
\label{tab:smallResults}
\resizebox{\textwidth}{!}{\begin{tabular}{|c|c|c|c|c|c|c|c|c|c|}
\hline
Data set                        & Algorithm & 0\%                  & 10\%                 & 20\%                 & 30\%                 & 40\%                 & 50\%                 & 60\%                 & 70\%                 \\ \hline
\multirow{2}{*}{RSMAS}          & Baseline  & \textbf{97.78 $\pm$ 0.94} & 93.50 $\pm$ 1.60          & 88.04 $\pm$ 3.42          & 82.59 $\pm$ 2.85          & 74.10 $\pm$ 3.73          & 63.91 $\pm$ 3.72          & 52.54 $\pm$ 4.15          & 38.83 $\pm$ 5.64          \\ \cline{2-10} 
                                & RAFNI     & 97.70 $\pm$ 1.39          & \textbf{96.76 $\pm$ 1.49} & \textbf{95.51 $\pm$ 2.16} & \textbf{92.09 $\pm$ 1.68} & \textbf{88.95 $\pm$ 2.48} & \textbf{79.13 $\pm$ 3.89} & \textbf{67.04 $\pm$ 4.42} & \textbf{53.73 $\pm$ 4.68} \\ \hline
\multirow{2}{*}{EILAT}          & Baseline  & \textbf{97.53 $\pm$ 1.51}          & 94.65 $\pm$ 2.09          & 89.67 $\pm$ 1.81          & 84.10 $\pm$ 2.40          & 75.96 $\pm$ 4.44          & 65.44 $\pm$ 4.24          & 53.30 $\pm$ 4.11          & 37.46 $\pm$ 5.18          \\ \cline{2-10} 
                                & RAFNI     & 97.41 $\pm$ 1.64 & \textbf{96.37 $\pm$ 1.57} & \textbf{95.80 $\pm$ 1.94} & \textbf{95.34 $\pm$ 2.04} & \textbf{93.60 $\pm$ 2.10} & \textbf{90.76 $\pm$ 2.77} & \textbf{86.72 $\pm$ 3.34} & \textbf{76.67 $\pm$ 5.57} \\ \hline
\multirow{2}{*}{StructureRSMAS} & Baseline  & 81.73 $\pm$ 4.42 & 80.23 $\pm$ 4.06          & 74.37 $\pm$ 5.00          & 70.87 $\pm$ 5.39          & 62.59 $\pm$ 5.05          & 51.60 $\pm$ 7.01          & 40.90 $\pm$ 6.16          & 32.98 $\pm$ 4.06          \\ \cline{2-10} 
                                & RAFNI     & \textbf{82.05 $\pm$ 3.78}          & \textbf{81.17 $\pm$ 3.63} & \textbf{77.40 $\pm$ 4.99} & \textbf{74.09 $\pm$ 5.35} & \textbf{68.43 $\pm$ 5.24} & \textbf{54.92 $\pm$ 6.22} & \textbf{46.57 $\pm$ 8.93} & \textbf{35.90 $\pm$ 6.04} \\ \hline
\end{tabular}}
\end{table*}

In Table~\ref{tab:smallResults}, we can observe the results with symmetric noise for the data sets RSMAS, EILAT and StructureRSMAS using RAFNI with ResNet50 as the backbone network, and the comparison with the backbone network alone as the baseline. The results are similar in all data sets as the noise increases, especially for RSMAS and EILAT. At 0\% of noise, the difference between the use of the RAFNI algorithm and the baseline is minimal. Then, as noise increases, this difference starts to increase. At 10\% of noise, RAFNI obtains 3.26\% more than the baseline for RSMAS and 1.72\% for EILAT. At 40\% of noise, this difference is 14.85\% for RSMAS and 17.64\% for EILAT. At 70\% of noise, the gain of using RAFNI is 14.9\% for RSMAS and 39.21\% for EILAT. For StructureRSMAS, these differences are lower: for example, at 40\%, the gain of using RAFNI is 5.84\%. However, RAFNI is still consistently better than the baseline at all levels of noise.

\begin{table}[]
\centering
\caption{5x5fcv mean $\pm$ std accuracy obtained for the data set COVIDGR1.0-SN using RAFNI with ResNet50 as backbone network and the backbone network alone, ResNet50, as baseline. The best results in each case are stressed in bold.}
\label{tab:covidsnResults}
\begin{tabular}{|c|c|c|}
\hline
Noise & Baseline    & RAFNI                \\ \hline
0\%   & 77.06 $\pm$ 3.47 & \textbf{78.20 $\pm$ 2.80} \\ \hline
10\%  & 73.46 $\pm$ 2.38 & \textbf{76.31 $\pm$ 2.44} \\ \hline
20\%  & 72.71 $\pm$ 4.74 & \textbf{75.91 $\pm$ 2.92} \\ \hline
30\%  & 68.14 $\pm$ 4.00 & \textbf{75.06 $\pm$ 3.94} \\ \hline
40\%  & 62.49 $\pm$ 2.87 & \textbf{72.77 $\pm$ 4.68} \\ \hline
50\%  & 55.34 $\pm$ 4.33 & \textbf{64.46 $\pm$ 5.68} \\ \hline
\end{tabular}
\end{table}

The results obtained for COVIDGR1.0-SN with pseudo-symmetric noise are shown in Table~\ref{tab:covidsnResults}. Here we only used levels of noise up until 50\% because this data set has only two classes. This data set has the advantage that it is a real-world data set, and it is more difficult to train (at 0\% noise level) than the other data sets:  ResNet50 obtains an accuracy of 77.06\% at 0\% noise. In addition, the noise we introduced in this data set is more realistic, so we can see how well the RAFNI algorithm behaves in a more real-life scenario. We can see that, at all noise levels, including 0\%, the results are better using RAFNI than using the baseline, with gains that generally increase as the noise level raises. At 10\% noise, RAFNI obtains 2.85\% more than the baseline. This gain is 6.92\% at 30\% noise and 9.12\% at 50\% noise.

\subsection{CIFAR}
\label{subsec:cifar}

We show the results we have obtained using CIFAR10 and CIFAR100 with symmetric noise and CIFAR10 with asymmetric noise.

\begin{table*}[]
\centering
\caption{Mean $\pm$ std accuracy obtained using CIFAR10 and CIFAR100 with symmetric noise and using the baseline network (ResNet50) and RAFNI with that network as the backbone network. The best results in each case are stressed in bold.}
\label{tab:CIFAR-sym}
\resizebox{\textwidth}{!}{\begin{tabular}{|c|c|c|c|c|c|c|c|c|}
\hline
Data set & \multicolumn{4}{c|}{CIFAR10}                                                                                                                             & \multicolumn{4}{c|}{CIFAR100}                                                                                                                            \\ \hline
Noise    & 0\%                  & 20\%                 & 40\%                 & 60\%                 & 0\%                  & 20\%                 & 40\%                 & 60\%                 \\ \hline
Baseline & 95.31 $\pm$ 0.03          & 85.36 $\pm$ 0.41          & 67.94 $\pm$ 0.50         & 44.87 $\pm$ 0.34          & \textbf{81.18 $\pm$ 0.26} & 70.57 $\pm$ 0.25          & 55.88 $\pm$ 0.45          & 37.64 $\pm$ 0.47          \\ \hline
RAFNI    & \textbf{95.48 $\pm$ 0.09} & \textbf{92.86 $\pm$ 0.32} & \textbf{89.84 $\pm$ 0.66} & \textbf{79.48 $\pm$ 1.23} & 80.68 $\pm$ 0.14          & \textbf{77.93 $\pm$ 0.11} & \textbf{73.01 $\pm$ 0.16} & \textbf{67.08 $\pm$ 0.45} \\ \hline
\end{tabular}}
\end{table*}

In Table~\ref{tab:CIFAR-sym} we can see the results for CIFAR10 and CIFAR100 using symmetric noise. RAFNI achieves better results in both data sets at all levels of noise except for CIFAR100 at 0\% of noise, where the baseline is slightly better. Similarly to what happened with the small datasets, the accuracy gain of using RAFNI increases as the noise level increases. For CIFAR10 we have a gain of 7.5\% at 20\% of noise and a gain of 34.61\% at 60\% of noise. For CIFAR100 these gains are 7.36\% and 29.44\% at 20\% and 60\% of noise, respectively.

\begin{table}[]
\centering
\caption{Mean $\pm$ std accuracy obtained using CIFAR10 with asymmetric noise and using the baseline network (ResNet50) and RAFNI with that network as the backbone network. The best results in each case are stressed in bold.}
\label{tab:CIFAR-asym}
\begin{tabular}{|c|c|c|c|c|}
\hline
Noise    & 0\%                  & 20\%                 & 30\%                 & 40\%                 \\ \hline
Baseline & 95.31 $\pm$ 0.03          & 89.27 $\pm$ 0.43          & 84.27 $\pm$ 0.30          & 78.10 $\pm$ 0.52          \\ \hline
RAFNI    & \textbf{95.48 $\pm$ 0.09} & \textbf{93.96 $\pm$ 0.15} & \textbf{92.96 $\pm$ 0.13} & \textbf{88.51 $\pm$ 0.54} \\ \hline
\end{tabular}
\end{table}

In Table~\ref{tab:CIFAR-asym} we show the results for CIFAR10 using asymmetric noise. In this scenario, RAFNI also achieves better results than the baseline at all levels of noise. The accuracy gain of using RAFNI increases when the noise increases, being 4.69\% at 20\% of noise, 8.69\% at 40\% of noise and 10.41\% at 60\% of noise.

\section{Comparison with state-of-the-art models}
\label{sec:comparison}

In this section, we compare our proposal, RAFNI, with some state-of-the-art models that do not used external information (like the noise rate): the loss correction approaches proposed in \cite{patrini2017making}, the robust function proposed in \cite{NEURIPS2019_8cd7775f}, the proposal in \cite{ma2018dimensionality}, and the one in \cite{arazo2019unsupervised}, which we described in Section~\ref{sec:background}.

The two proposed methods in \cite{patrini2017making} suppose that the noise rate is known, but the authors incorporated a mechanism to estimate it in the usual case that it is not known, so we used both of their approaches using this estimation (called estimated forward and estimated backwards).

To make the comparison we used CIFAR10, with symmetric and asymmetric noise, and CIFAR100 with symmetric noise, which are the benchmarks that most of the papers used in the literature.

To make a fair comparison, we used the same experimental frameworks as the other papers whenever possible, that is, we changed our framework to use the same backbone neural network, the same number of training epochs, optimizer, the same learning rate scheduler, data augmentation, etc., as the model we are comparing our algorithm to. In each case, we used the best hyperparameters reported in each paper for each scenario, except for the number of epochs. Due to time restrictions, if the original number of epochs used by the authors exceeds 120 for CIFAR10 and 150 for CIFAR100, we changed them to use 120 epochs for CIFAR10 and 150 epochs for CIFAR100 (accordingly, we also train RAFNI for the same number of epochs in each case). The authors in \cite{NEURIPS2019_8cd7775f} only used CIFAR100 under symmetric noise, so we only had the best values for the two temperatures for this case. For the other two scenarios (CIFAR10 with symmetric noise and with asymmetric noise), we evaluated different values in the range the authors gave for each hyperparameter and selected the best ones for each scenario and level of noise using the same validation set we used to search for the best hyperparameters for RAFNI. For this proposal, we use ResNet50 in all scenarios, since their method can be used with any CNN.

The authors in \cite{arazo2019unsupervised} used an original implementation of pre-activation ResNet18 in PyTorch. Unfortunately, we were not able to replicate that network with our algorithm. As a result, we compared our results using RAFNI with our experimental scheme (ResNet50 with fine-tuning, 10 and 15 epochs for CIFAR10 and CIFAR100 respectively) with their model using their experimental scheme (ResNet18, 300 epochs in both cases, data augmentation and learning rate scheduler). In their paper, they also compared their method with other proposals in the literature using different backbone networks.

Finally, we used the same data sets as with our algorithm where it was possible (for the proposals made by \cite{NEURIPS2019_8cd7775f} and  \cite{ma2018dimensionality}), and the given ones in the rest, making sure that the noise injection was the same as it was in our data sets. We argue that, since the noise level is the same, it is introduced randomly in all the cases, and the test sets are also the same as they are predefined, we can safely compare the algorithms.

We also performed a Wilcoxon Rank-Sum test to check if the differences in the results were significant in each case. Since we repeated each experiment five times, we used all five accuracies for each data set and level of noise to perform the Wilcoxon test, instead of using the mean.

\begin{table*}[]
\centering
\caption{Comparison between RAFNI and the two methods from Patrini et al \cite{patrini2017making}, using pre-activation ResNet32 for CIFAR10 and pre-activation ResNet44 for CIFAR100 in the three approaches. The best results are stressed in bold.}
\label{tab:compPatrini}
\resizebox{\textwidth}{!}{\begin{tabular}{|c|cccccc|ccc|}
\hline
\multirow{3}{*}{} & \multicolumn{6}{c|}{CIFAR10}                                                                                                                                                                                                                     & \multicolumn{3}{c|}{CIFAR100}                                                                                \\ \cline{2-10} 
                  & \multicolumn{3}{c|}{Symmetric noise}                                                                                              & \multicolumn{3}{c|}{Asymmetric noise}                                                                        & \multicolumn{3}{c|}{Symmetric noise}                                                                         \\ \cline{2-10} 
                  & \multicolumn{1}{c|}{20\%}                 & \multicolumn{1}{c|}{40\%}                 & \multicolumn{1}{c|}{60\%}                 & \multicolumn{1}{c|}{20\%}                 & \multicolumn{1}{c|}{30\%}                 & 40\%                 & \multicolumn{1}{c|}{20\%}                 & \multicolumn{1}{c|}{40\%}                 & 60\%                 \\ \hline
Est. Forward      & \multicolumn{1}{c|}{\textbf{88.46 $\pm$ 0.22}} & \multicolumn{1}{c|}{\textbf{84.54 $\pm$ 0.40}} & \multicolumn{1}{c|}{79.32 $\pm$ 0.23}          & \multicolumn{1}{c|}{\textbf{89.89 $\pm$ 0.15}} & \multicolumn{1}{c|}{\textbf{89.32 $\pm$ 0.28}} & 87.09 $\pm$ 1.55          & \multicolumn{1}{c|}{\textbf{62.11 $\pm$ 2.57}} & \multicolumn{1}{c|}{48.72 $\pm$ 0.84}          & 33.41 $\pm$ 0.87          \\ \hline
Est. Backwards     & \multicolumn{1}{c|}{84.40 $\pm$ 0.22}          & \multicolumn{1}{c|}{79.12 $\pm$ 0.45}          & \multicolumn{1}{c|}{64.00 $\pm$ 1.67}          & \multicolumn{1}{c|}{86.33 $\pm$ 0.55}          & \multicolumn{1}{c|}{81.71 $\pm$ 3.06}          & 72.10 $\pm$ 5.66          & \multicolumn{1}{c|}{60.24 $\pm$ 3.36}          & \multicolumn{1}{c|}{--}                   & --                   \\ \hline
RAFNI             & \multicolumn{1}{c|}{87.58 $\pm$ 0.19}          & \multicolumn{1}{c|}{84.38$\pm$  0.56}          & \multicolumn{1}{c|}{\textbf{79.34 $\pm$ 0.51}} & \multicolumn{1}{c|}{88.79 $\pm$ 0.19}          & \multicolumn{1}{c|}{87.33 $\pm$ 0.25}          & \textbf{87.30 $\pm$ 1.12} & \multicolumn{1}{c|}{61.49 $\pm$ 1.57}          & \multicolumn{1}{c|}{\textbf{55.20 $\pm$ 1.18}} & \textbf{45.31 $\pm$ 0.96} \\ \hline
\end{tabular}}
\end{table*}

In Table~\ref{tab:compPatrini} we show the accuracy obtained with the two approaches proposed in \cite{patrini2017making} and our algorithm using the same backbone network and experimental scheme as the one used in \cite{patrini2017making}. In the comparison with the estimated backward algorithm, we can see that RAFNI performs better in both data sets at all levels and types of noise. It is interesting that when classifying CIFAR100 with a noise rate of 40\% and more, the estimated backward algorithm does not finish the training. Here, the Wilcoxon-Rank-Sum test obtains that there are significant differences with p-value $9.5\times 10^{-7}$.

When comparing RAFNI with the estimated forward algorithm, we can see that there is less difference between them, especially in CIFAR10, where the differences are usually less than 1\% using both types of noise, on average. However, when classifying CIFAR100, RAFNI outperforms the estimated forward algorithm when the noise rate is 40\% or more. In particular, RAFNI obtains a gain in accuracy of 6.48\% at 40\% noise and 11.9\% at 60\% noise. Using the Wilcoxon Rank-Sum test, we can say that RAFNI performs better with significant differences (p-value $3.6\times 10^{-5}$.

\begin{table*}[]
\centering
\caption{Comparison between RAFNI and the D2L method \cite{ma2018dimensionality}, using their original 8-layer CNN for CIFAR10 and pre-activation ResNet44 for CIFAR100 in both approaches. The best results are stressed in bold}
\label{tab:compD2L}
\resizebox{\textwidth}{!}{\begin{tabular}{|c|cccccc|ccc|}
\hline
\multirow{3}{*}{} & \multicolumn{6}{c|}{CIFAR10}                                                                                                                                                                                                                     & \multicolumn{3}{c|}{CIFAR100}                                                                                \\ \cline{2-10} 
                  & \multicolumn{3}{c|}{Symmetric noise}                                                                                              & \multicolumn{3}{c|}{Asymmetric noise}                                                                        & \multicolumn{3}{c|}{Symmetric noise}                                                                         \\ \cline{2-10} 
                  & \multicolumn{1}{c|}{20\%}                 & \multicolumn{1}{c|}{40\%}                 & \multicolumn{1}{c|}{60\%}                 & \multicolumn{1}{c|}{20\%}                 & \multicolumn{1}{c|}{30\%}                 & 40\%                 & \multicolumn{1}{c|}{20\%}                 & \multicolumn{1}{c|}{40\%}                 & 60\%                 \\ \hline
D2L               & \multicolumn{1}{c|}{86.43 $\pm$ 0.15}          & \multicolumn{1}{c|}{\textbf{84.08 $\pm$ 0.30}} & \multicolumn{1}{c|}{\textbf{78.32 $\pm$ 0.48}} & \multicolumn{1}{c|}{86.75 $\pm$ 0.14}          & \multicolumn{1}{c|}{85.27 $\pm$ 0.27}          & 82.51 $\pm$ 0.34          & \multicolumn{1}{c|}{59.18 $\pm$ 0.30}          & \multicolumn{1}{c|}{28.77 $\pm$ 8.52}          & 4.34 $\pm$ 1.90           \\ \hline
RAFNI             & \multicolumn{1}{c|}{\textbf{87.91 $\pm$ 0.19}} & \multicolumn{1}{c|}{83.01 $\pm$ 0.81}           & \multicolumn{1}{c|}{76.37 $\pm$ 0.15}          & \multicolumn{1}{c|}{\textbf{89.04 $\pm$ 0.13}} & \multicolumn{1}{c|}{\textbf{88.49 $\pm$ 0.19}} & \textbf{86.57 $\pm$ 0.17} & \multicolumn{1}{c|}{\textbf{61.49 $\pm$ 1.57}} & \multicolumn{1}{c|}{\textbf{55.20 $\pm$ 1.18}} & \textbf{45.31 $\pm$ 0.96} \\ \hline
\end{tabular}}
\end{table*}

The results we obtained for the D2L algorithm \cite{ma2018dimensionality} and our algorithm using the same experimental scheme can be seen in Table~\ref{tab:compD2L}. We can see that for CIFAR10 with symmetric noise there is not much difference between the accuracies obtained by D2L and RAFNI, with D2L being slightly better at 60\% noise with a difference of 1.95\% on average. But when introducing asymmetric noise, RAFNI obtained better results at all noise rates, with a difference in the accuracy of 4.06\% on average at 40\% noise. The biggest difference between these two methods, however, can be seen when classifying CIFAR100, where RAFNI outperforms D2L, especially as noise increases, with a difference of 26.43\% at 40\% noise and 40.97\% at 60\% noise. The Wilcoxon Rank-Sum test obtains significant differences with p-value $1.31\times 10^{-5}$.

\begin{table*}[]
\centering
\caption{Comparison between RAFNI and the BiTempered method \cite{NEURIPS2019_8cd7775f}, using ResNet50 in both approaches. The best results are stressed in bold}
\label{tab:compBiT}
\resizebox{\textwidth}{!}{\begin{tabular}{|c|cccccc|ccc|}
\hline
\multirow{3}{*}{} & \multicolumn{6}{c|}{CIFAR10}                                                                                                                                                                                                                     & \multicolumn{3}{c|}{CIFAR100}                                                                                \\ \cline{2-10} 
                  & \multicolumn{3}{c|}{Symmetric noise}                                                                                              & \multicolumn{3}{c|}{Asymmetric noise}                                                                        & \multicolumn{3}{c|}{Symmetric noise}                                                                         \\ \cline{2-10} 
                  & \multicolumn{1}{c|}{20\%}                 & \multicolumn{1}{c|}{40\%}                 & \multicolumn{1}{c|}{60\%}                 & \multicolumn{1}{c|}{20\%}                 & \multicolumn{1}{c|}{30\%}                 & 40\%                 & \multicolumn{1}{c|}{20\%}                 & \multicolumn{1}{c|}{40\%}                 & 60\%                 \\ \hline
BiTempered        & \multicolumn{1}{c|}{89.33 $\pm$ 0.22}          & \multicolumn{1}{c|}{76.14 $\pm$ 1.12}          & \multicolumn{1}{c|}{54.14 $\pm$ 1.37}          & \multicolumn{1}{c|}{88.82 $\pm$ 0.73}          & \multicolumn{1}{c|}{84.09 $\pm$ 0.63}          & 78.07 $\pm$ 0.54          & \multicolumn{1}{c|}{76.51 $\pm$ 0.22}          & \multicolumn{1}{c|}{71.31 $\pm$ 0.32}          & 64.47 $\pm$ 0.43          \\ \hline
RAFNI             & \multicolumn{1}{c|}{\textbf{92.86 $\pm$ 0.32}} & \multicolumn{1}{c|}{\textbf{89.84 $\pm$ 0.66}} & \multicolumn{1}{c|}{\textbf{79.48 $\pm$ 1.23}} & \multicolumn{1}{c|}{\textbf{93.96 $\pm$ 0.15}} & \multicolumn{1}{c|}{\textbf{92.96 $\pm$ 0.13}} & \textbf{88.51 $\pm$ 0.54} & \multicolumn{1}{c|}{\textbf{77.93 $\pm$ 0.11}} & \multicolumn{1}{c|}{\textbf{73.01 $\pm$ 0.16}} & \textbf{67.08 $\pm$ 0.45} \\ \hline
\end{tabular}}
\end{table*}

The accuracies we obtained using the BiTempered method and RAFNI, both of them using ResNet50 as the backbone network, can be seen in Table~\ref{tab:compBiT}. Here, RAFNI obtains, on average, better results in both data sets at all noise rates and noise types. The Wilcoxon Rank-Sum obtains significant differences with p-value $1.77\times 10^{-8}$. The biggest differences occur for CIFAR10 with symmetric noise, where RAFNI outperforms Bi-Tempered by 13.7\% at 40\% noise and by 25.34\% at 60\% noise.

\begin{table*}[]
\centering
\caption{Comparison between RAFNI, using ResNet50, and the method from Arazo et al \cite{arazo2019unsupervised}, using pre-activation ResNet18. The best results are stressed in bold}
\label{tab:compArazo}
\resizebox{\textwidth}{!}{\begin{tabular}{|c|cccccc|ccc|}
\hline
\multirow{3}{*}{} & \multicolumn{6}{c|}{CIFAR10}                                                                                                                                                                                                                     & \multicolumn{3}{c|}{CIFAR100}                                                                                \\ \cline{2-10} 
                  & \multicolumn{3}{c|}{Symmetric noise}                                                                                              & \multicolumn{3}{c|}{Asymmetric noise}                                                                        & \multicolumn{3}{c|}{Symmetric noise}                                                                         \\ \cline{2-10} 
                  & \multicolumn{1}{c|}{20\%}                 & \multicolumn{1}{c|}{40\%}                 & \multicolumn{1}{c|}{60\%}                 & \multicolumn{1}{c|}{20\%}                 & \multicolumn{1}{c|}{30\%}                 & 40\%                 & \multicolumn{1}{c|}{20\%}                 & \multicolumn{1}{c|}{40\%}                 & 60\%                 \\ \hline
Arazo et al \cite{arazo2019unsupervised}      & \multicolumn{1}{c|}{\textbf{93.54 $\pm$ 0.30}} & \multicolumn{1}{c|}{\textbf{92.45 $\pm$ 0.18}} & \multicolumn{1}{c|}{\textbf{89.34 $\pm$ 0.20}} & \multicolumn{1}{c|}{87.60 $\pm$ 3.15}          & \multicolumn{1}{c|}{80.68 $\pm$ 6.46}          & 64.49 $\pm$ 27.52         & \multicolumn{1}{c|}{69.29 $\pm$ 0.16}          & \multicolumn{1}{c|}{63.72 $\pm$ 0.19}          & 54.83 $\pm$ 0.45          \\ \hline
RAFNI             & \multicolumn{1}{c|}{92.86 $\pm$ 0.32}          & \multicolumn{1}{c|}{89.84 $\pm$ 0.66}          & \multicolumn{1}{c|}{79.48 $\pm$ 1.23}          & \multicolumn{1}{c|}{\textbf{93.96 $\pm$ 0.15}} & \multicolumn{1}{c|}{\textbf{92.96 $\pm$ 0.13}} & \textbf{88.51 $\pm$ 0.54} & \multicolumn{1}{c|}{\textbf{77.93 $\pm$ 0.11}} & \multicolumn{1}{c|}{\textbf{73.01 $\pm$ 0.16}} & \textbf{67.08 $\pm$ 0.45} \\ \hline
\end{tabular}}
\end{table*}

Finally, the accuracies we obtained with the algorithm proposed by Arazo et al in \cite{arazo2019unsupervised} and RAFNI, both of them using their original experimental schemes, can be seen in Table~\ref{tab:compArazo}. This is the case where we can see the most discrepancies depending on the data set that is being classified but also depending on the type of noise. For CIFAR10 with symmetric noise, which is easier to classify than CIFAR100, the algorithm from Arazo et al is better than RAFNI, obtaining 9.86\% more at 60\% noise, on average, though the difference at 20\% is considerably lower. However, if we introduce asymmetric noise on CIFAR10, which is a more complicated and realistic type of noise, RAFNI obtains better results. The differences increase as the noise rate increases, being 24.02\% at 40\% noise. Now, for CIFAR100 with symmetric noise, RAFNI again outperforms the algorithm proposed by Arazo et al, and the differences in accuracy are present even at low percentages of noise, similar to what happened for CIFAR10 with asymmetric noise: RAFNI obtains 8.64\% more accuracy, on average, at 20\% noise, 9.29\% at 40\% noise and 12.25\% at 60\% noise. When we used the Wilcoxon Rank-Sum test on all results, we obtained that RAFNI obtained significant differences with p-value $1.15\times 10^{-7}$.

\section{Comparison with an approach that suppose the noise rate is known}
\label{sec:compSELFIE}

In this section, we compare our algorithm, which assumes no known external information, with the SELFIE algorithm, which assumes the noise rate of the data set is known. It is important to note that, in real scenarios, knowing the noise rate of the data set is not usual. We want to estimate the advantage of using this unusual information in specific models, so that we can quantify the potential loss of not having that kind of knowledge available. Thus, RAFNI is expected to perform worse than SELFIE since RAFNI does not assume extra information, but we want to see in which cases it is competitive.

To make this comparison, we used RAFNI with the same experimental scheme that SELFIE: DenseNet-25-12 as the backbone network, with no data augmentation and the same learning rate scheduler. We train both algorithms for 120 and 150 epochs for CIFAR10 and CIFAR100, respectively, due to time restrictions. For SELFIE, we had to implement the matrix of asymmetric noise as given in \cite{patrini2017making}, to use the same injection of noise in the data sets used in both algorithms.

\begin{table*}[]
\centering
\caption{Comparison between RAFNI and SELFIE \cite{song2019selfie}, using DenseNet-25-12. The best results are stressed in bold}
\label{tab:compSELFIE}
\resizebox{\textwidth}{!}{\begin{tabular}{|c|cccccc|ccc|}
\hline
\multirow{3}{*}{} & \multicolumn{6}{c|}{CIFAR10}                                                                                                                                                                                                                     & \multicolumn{3}{c|}{CIFAR100}                                                                                \\ \cline{2-10} 
                  & \multicolumn{3}{c|}{Symmetric noise}                                                                                              & \multicolumn{3}{c|}{Asymmetric noise}                                                                        & \multicolumn{3}{c|}{Symmetric noise}                                                                         \\ \cline{2-10} 
                  & \multicolumn{1}{c|}{20\%}                 & \multicolumn{1}{c|}{40\%}                 & \multicolumn{1}{c|}{60\%}                 & \multicolumn{1}{c|}{20\%}                 & \multicolumn{1}{c|}{30\%}                 & 40\%                 & \multicolumn{1}{c|}{20\%}                 & \multicolumn{1}{c|}{40\%}                 & 60\%                 \\ \hline
SELFIE            & \multicolumn{1}{c|}{\textbf{88.04 $\pm$ 0.21}} & \multicolumn{1}{c|}{\textbf{85.26 $\pm$ 0.30}} & \multicolumn{1}{c|}{\textbf{77.29 $\pm$ 0.45}} & \multicolumn{1}{c|}{\textbf{87.90 $\pm$ 0.33}} & \multicolumn{1}{c|}{\textbf{84.01 $\pm$ 0.24}} & 66.57 $\pm$ 0.68          & \multicolumn{1}{c|}{\textbf{63.48 $\pm$ 0.67}} & \multicolumn{1}{c|}{\textbf{60.00 $\pm$ 0.40}} & \textbf{52.62 $\pm$ 0.49} \\ \hline
RAFNI             & \multicolumn{1}{c|}{84.00 $\pm$ 0.23}          & \multicolumn{1}{c|}{78.62 $\pm$ 0.84}          & \multicolumn{1}{c|}{68.27 $\pm$ 0.31}          & \multicolumn{1}{c|}{84.48 $\pm$ 0.70}          & \multicolumn{1}{c|}{82.51 $\pm$ 0.65}          & \textbf{78.19 $\pm$ 1.83} & \multicolumn{1}{c|}{56.08 $\pm$ 0.33}          & \multicolumn{1}{c|}{48.00 $\pm$ 0.83}             & 31.91 $\pm$ 1.38             \\ \hline
\end{tabular}}
\end{table*}

The results we obtained for both algorithms are shown in Table~\ref{tab:compSELFIE}. As expected, SELFIE obtained better results for both data sets and types of noise. Nonetheless, for the CIFAR10 data set RAFNI performs well, obtaining similar results at low noise rates. This also happened when using asymmetric noise, which is interesting since this type of noise is more realistic and difficult to deal with than symmetric noise. In fact, RAFNI outperforms SELFIE at 40\% noise by more than 11\%.

However, in general, if the information about the noise rate is known, it is better to use an algorithm that uses that information. We did not suppose this type of information was known in our algorithm as we wanted to make it applicable in all cases.

\section{Analysing the effectiveness of the RAFNI mechanisms}
\label{sec:effectiveness}

In this section, we analyse how well RAFNI removes and relabels instances using two data sets as samples: EILAT and COVIDGR1.0-SN. In particular, we take a look at a) how many instances were removed, and from those, we check how many of them were noisy; and b) how many instances were relabelled, and from those, we check how many of them were noisy and how many were correctly classified by the algorithm to their original class.

For both data sets, we analysed how well RAFNI behaves at every level of noise tested in Section~\ref{sec:results}. We used one five-fold cross-validation repeated five times for both data sets, so the results we give here are the total results, that is, for the total number of removals, for example, we sum all the removals in the five training sets. The results for EILAT can be seen in Table~\ref{tab:anaEILAT} and the results for COVIDGR1.0-SN can be seen in Table~\ref{tab:anaCOVID}. In both tables we show 1) the percentage of good removals, that is, instances that were noisy and RAFNI removed from the training set; 2) the total number of instances that RAFNI removed during training; 3) the percentage of good changes, that is, instances that were noisy and RAFNI changed to their original clean class; 4) the total number of instances that RAFNI changed from one class to another during training. Since EILAT has more than two classes, in its case we also show the percentage of noisy changes, that is, instances that were noisy and RAFNI changed to another class, but not their original class.

In both cases, we can see that RAFNI does a good job both removing and changing instances to their original class. In the case of the COVIDGR1.0-SN data set, the percentage of good changes is lower than with EILAT, but this is to be expected since the type of noise introduced in the COVIDGR1.0-SN data set is more difficult and realistic than the symmetric noise introduced in EILAT. Even in that case, RAFNI removes instances that were noisy with an accuracy above 92\% at all levels of noise for COVIDGR1.0-SN. On the other hand, RAFNI changes instances to their original class with precision above 95\% for EILAT, except at 70\% of noise, when it descends to 81.81\%.  This shows that RAFNI is capable to detect the noisy instances and either remove them or change them to their original class with a high precision, which improves the learning, as we have seen in Section~\ref{sec:results}. The total number of removals and changes tends to increase as the noise level increases in both data sets, as would be expected since the number of noisy instances is increasing, so this is another sign that the algorithm is behaving well.

\begin{table*}[]
\centering
\caption{Analysis of the instances that the RAFNI algorithm removed and changed from one class to another during the training of the EILAT data set.}
\label{tab:anaEILAT}
\begin{tabular}{|c|c|c|c|c|c|c|c|}
\hline
Noise                    & 10\%    & 20\%    & 30\%    & 40\%    & 50\%    & 60\%    & 70\%    \\ \hline
\% good removals         & 75.80\% & 73.58\% & 83.58\% & 88.53\% & 85.66\% & 84.80\% & 82.28\% \\ \hline
Total number of removals & 2595    & 4969    & 6706    & 9039    & 18519   & 13423   & 15313   \\ \hline
\% good changes          & 100\%   & 99.75\% & 99.22\% & 98.98\% & 98.36\% & 95.63\% & 81.81\% \\ \hline
\% noisy changes         & 0\%     & 0\%     & 0.10\%  & 0.44\%  & 0.44\%  & 2.15\%  & 10.29\% \\ \hline
Total number of changes  & 156     & 809     & 1022    & 683     & 3176    & 1486    & 2303    \\ \hline
\end{tabular}
\end{table*}

\begin{table}[]
\centering
\caption{Analysis of the instances that the RAFNI algorithm removed and changed from one class to another during the training of the COVIDGR1.0-SN data set.}
\label{tab:anaCOVID}
\resizebox{\columnwidth}{!}{\begin{tabular}{|c|c|c|c|c|c|}
\hline
Noise                    & 10\%    & 20\%    & 30\%    & 40\%    & 50\%    \\ \hline
\% good removals         & 94.64\% & 94.15\% & 94.16\% & 92.92\% & 93.27\% \\ \hline
Total number of removals & 1866    & 4719    & 4487    & 6301    & 7282    \\ \hline
\% good changes          & 40.61\% & 68.39\% & 70.88\% & 74.58\% & 70.13\% \\ \hline
Total number of changes  & 1091    & 291     & 443     & 535     & 385     \\ \hline
\end{tabular}}
\end{table}

\section{Conclusions}
\label{sec:conclusions}

In this paper, we proposed an algorithm, called RAFNI, that can filter and relabel noisy instances during the training process of any convolutional neural network using the predictions and loss values the network gives the instances of the training set. This progressive cleaning of the training set allows the network to improve its generalisation at the end of the training process, improving the results the CNN has on its own. In addition, RAFNI has the advantage that it can be used with any CNN as the backbone network and that transfer learning and data augmentation can be easily applied. It also does not use prior information that is usually not known, like the noise matrix or the noise rate. In addition, it works well even when there is no introduced noise in the data set, so it is safe to use when we do not know the noise rate of a data set. We also made the code available so it is easier to use it.

Developing algorithms that can allow deep neural networks to perform better under label noise is an important task since label noise is a common problem in real-world scenarios and it negatively affects the performance of the networks. We believe that our proposal is a great solution to this problem: it can be easily fine-tuned to every data set, it allows to be used with any CNN, and it allows the use of transfer learning and data augmentation. We proved its potential using various data sets with different characteristics and using three different types of label noise. Finally, we also compared it with several state-of-the-art algorithms, improving their results.

\section*{Acknowledgements}
This publication was supported by the project with reference SOMM17/6110/UGR, granted by the Andalusian Consejería de Conocimiento, Investigación y  Universidades and European Regional Development Funds (ERDF). This work was also supported by project PID2020-119478GB-I00 granted by Ministerio de Ciencia, Innovación y Univesidades, and project P18-FR-4961 by Proyectos
I+D+i Junta de Andalucia 2018. Anabel Gómez-Ríos was supported by the FPU Programme FPU16/04765 by Ministerio de Educación, Cultura y Deporte. 



 
\bibliography{mybibfile}

\begin{thebibliography}{10}
\providecommand{\url}[1]{#1}
\csname url@samestyle\endcsname
\providecommand{\newblock}{\relax}
\providecommand{\bibinfo}[2]{#2}
\providecommand{\BIBentrySTDinterwordspacing}{\spaceskip=0pt\relax}
\providecommand{\BIBentryALTinterwordstretchfactor}{4}
\providecommand{\BIBentryALTinterwordspacing}{\spaceskip=\fontdimen2\font plus
\BIBentryALTinterwordstretchfactor\fontdimen3\font minus
  \fontdimen4\font\relax}
\providecommand{\BIBforeignlanguage}[2]{{%
\expandafter\ifx\csname l@#1\endcsname\relax
\typeout{** WARNING: IEEEtran.bst: No hyphenation pattern has been}%
\typeout{** loaded for the language `#1'. Using the pattern for}%
\typeout{** the default language instead.}%
\else
\language=\csname l@#1\endcsname
\fi
#2}}
\providecommand{\BIBdecl}{\relax}
\BIBdecl

\bibitem{krizhevsky2012imagenet}
A.~Krizhevsky, I.~Sutskever, and G.~E. Hinton, ``Imagenet classification with
  deep convolutional neural networks,'' \emph{Advances in neural information
  processing systems}, vol.~25, pp. 1097--1105, 2012.

\bibitem{GOMEZRIOS2019315}
\BIBentryALTinterwordspacing
A.~Gómez-Ríos, S.~Tabik, J.~Luengo, A.~Shihavuddin, B.~Krawczyk, and
  F.~Herrera, ``Towards highly accurate coral texture images classification
  using deep convolutional neural networks and data augmentation,''
  \emph{Expert Systems with Applications}, vol. 118, pp. 315--328, Mar 2019.
  [Online]. Available:
  \url{https://linkinghub.elsevier.com/retrieve/pii/S0957417418306523}
\BIBentrySTDinterwordspacing

\bibitem{olmos2018automatic}
\BIBentryALTinterwordspacing
R.~Olmos, S.~Tabik, and F.~Herrera, ``Automatic handgun detection alarm in
  videos using deep learning,'' \emph{Neurocomputing}, vol. 275, pp. 66--72,
  Jan 2018. [Online]. Available:
  \url{https://linkinghub.elsevier.com/retrieve/pii/S0925231217308196}
\BIBentrySTDinterwordspacing

\bibitem{song2020learning}
H.~Song, M.~Kim, D.~Park, Y.~Shin, and J.-G. Lee, ``Learning from noisy labels
  with deep neural networks: A survey,'' \emph{arXiv preprint
  arXiv:2007.08199}, 2020.

\bibitem{xiao2015learning}
T.~Xiao, T.~Xia, Y.~Yang, C.~Huang, and X.~Wang, ``Learning from massive noisy
  labeled data for image classification,'' in \emph{Proceedings of the IEEE
  conference on computer vision and pattern recognition}, 2015, pp. 2691--2699.

\bibitem{lee2017cleannet}
K.-H. Lee, X.~He, L.~Zhang, and L.~Yang, ``Cleannet: Transfer learning for
  scalable image classifier training with label noise,'' in \emph{Proceedings
  of the IEEE Conference on Computer Vision and Pattern Recognition ({CVPR})},
  2018.

\bibitem{6685834}
\BIBentryALTinterwordspacing
B.~Frenay and M.~Verleysen, ``Classification in the presence of label noise: A
  survey,'' \emph{IEEE Transactions on Neural Networks and Learning Systems},
  vol.~25, no.~5, pp. 845--869, May 2014. [Online]. Available:
  \url{http://ieeexplore.ieee.org/document/6685834/}
\BIBentrySTDinterwordspacing

\bibitem{li2017webvision}
W.~Li, L.~Wang, W.~Li, E.~Agustsson, and L.~Van~Gool, ``Webvision database:
  Visual learning and understanding from web data,'' \emph{arXiv preprint
  arXiv:1708.02862}, 2017.

\bibitem{song2019selfie}
H.~Song, M.~Kim, and J.-G. Lee, ``Selfie: Refurbishing unclean samples for
  robust deep learning,'' in \emph{International Conference on Machine
  Learning}.\hskip 1em plus 0.5em minus 0.4em\relax PMLR, 2019, pp. 5907--5915.

\bibitem{zhang2021understanding}
\BIBentryALTinterwordspacing
C.~Zhang, S.~Bengio, M.~Hardt, B.~Recht, and O.~Vinyals, ``Understanding deep
  learning (still) requires rethinking generalization,'' \emph{Communications
  of the ACM}, vol.~64, no.~3, pp. 107--115, Mar 2021. [Online]. Available:
  \url{https://dl.acm.org/doi/10.1145/3446776}
\BIBentrySTDinterwordspacing

\bibitem{patrini2017making}
G.~Patrini, A.~Rozza, A.~Krishna~Menon, R.~Nock, and L.~Qu, ``Making deep
  neural networks robust to label noise: A loss correction approach,'' in
  \emph{Proceedings of the IEEE Conference on Computer Vision and Pattern
  Recognition}, 2017, pp. 1944--1952.

\bibitem{ma2018dimensionality}
X.~Ma, Y.~Wang, M.~E. Houle, S.~Zhou, S.~Erfani, S.~Xia, S.~Wijewickrema, and
  J.~Bailey, ``Dimensionality-driven learning with noisy labels,'' in
  \emph{International Conference on Machine Learning}.\hskip 1em plus 0.5em
  minus 0.4em\relax PMLR, 2018, pp. 3355--3364.

\bibitem{jiang2018mentornet}
L.~Jiang, Z.~Zhou, T.~Leung, L.-J. Li, and L.~Fei-Fei, ``Mentornet: Learning
  data-driven curriculum for very deep neural networks on corrupted labels,''
  in \emph{International Conference on Machine Learning}.\hskip 1em plus 0.5em
  minus 0.4em\relax PMLR, 2018, pp. 2304--2313.

\bibitem{wang2018iterative}
Y.~Wang, W.~Liu, X.~Ma, J.~Bailey, H.~Zha, L.~Song, and S.-T. Xia, ``Iterative
  learning with open-set noisy labels,'' in \emph{Proceedings of the IEEE
  conference on computer vision and pattern recognition}, 2018, pp. 8688--8696.

\bibitem{10.5555/3326943.3327112}
G.~Song and W.~Chai, ``Collaborative learning for deep neural networks,'' in
  \emph{Proceedings of the 32nd International Conference on Neural Information
  Processing Systems}, ser. NIPS'18.\hskip 1em plus 0.5em minus 0.4em\relax
  Curran Associates Inc., 2018, p. 1837–1846.

\bibitem{jindal2016learning}
\BIBentryALTinterwordspacing
I.~Jindal, M.~Nokleby, and X.~Chen, ``Learning deep networks from noisy labels
  with dropout regularization.''\hskip 1em plus 0.5em minus 0.4em\relax IEEE,
  Dec 2016, pp. 967--972. [Online]. Available:
  \url{http://ieeexplore.ieee.org/document/7837934/}
\BIBentrySTDinterwordspacing

\bibitem{NEURIPS2019_8cd7775f}
E.~Amid, M.~K.~K. Warmuth, R.~Anil, and T.~Koren, ``Robust bi-tempered logistic
  loss based on bregman divergences,'' in \emph{Advances in Neural Information
  Processing Systems}, H.~Wallach, H.~Larochelle, A.~Beygelzimer,
  F.~d\textquotesingle Alch\'{e}-Buc, E.~Fox, and R.~Garnett, Eds.,
  vol.~32.\hskip 1em plus 0.5em minus 0.4em\relax Curran Associates, Inc.,
  2019.

\bibitem{ghosh2017robust}
A.~Ghosh, H.~Kumar, and P.~Sastry, ``Robust loss functions under label noise
  for deep neural networks,'' in \emph{Proceedings of the AAAI Conference on
  Artificial Intelligence}, vol.~31, no.~1, 2017.

\bibitem{zhang2018generalized}
Z.~Zhang and M.~R. Sabuncu, ``Generalized cross entropy loss for training deep
  neural networks with noisy labels,'' in \emph{32nd Conference on Neural
  Information Processing Systems (NeurIPS)}, 2018.

\bibitem{yi2019probabilistic}
K.~Yi and J.~Wu, ``Probabilistic end-to-end noise correction for learning with
  noisy labels,'' in \emph{Proceedings of the IEEE/CVF Conference on Computer
  Vision and Pattern Recognition}, 2019, pp. 7017--7025.

\bibitem{arazo2019unsupervised}
E.~Arazo, D.~Ortego, P.~Albert, N.~O’Connor, and K.~McGuinness,
  ``Unsupervised label noise modeling and loss correction,'' in
  \emph{International conference on machine learning}.\hskip 1em plus 0.5em
  minus 0.4em\relax PMLR, 2019, pp. 312--321.

\bibitem{c51d68a3106242f08ed001d0c46320b3}
S.~Sukhbaatar, J.~Bruna, M.~Paluri, L.~Bourdev, and R.~Fergus, ``Training
  convolutional networks with noisy labels,'' Jan. 2015, 3rd International
  Conference on Learning Representations, ICLR 2015.

\bibitem{reed2014training}
S.~Reed, H.~Lee, D.~Anguelov, C.~Szegedy, D.~Erhan, and A.~Rabinovich,
  ``Training deep neural networks on noisy labels with bootstrapping,''
  \emph{arXiv preprint arXiv:1412.6596}, 2014.

\bibitem{zhang2017mixup}
H.~Zhang, M.~Cisse, Y.~N. Dauphin, and D.~Lopez-Paz, ``mixup: Beyond empirical
  risk minimization,'' \emph{arXiv preprint arXiv:1710.09412}, 2017.

\bibitem{coralDataset}
A.~Shihavuddin, ``Coral reef dataset, mendeley data, v2,''
  \url{https://data.mendeley.com/datasets/86y667257h/2}, accesed on:
  06-04-2021.

\bibitem{GOMEZRIOS2019104891}
\BIBentryALTinterwordspacing
A.~Gómez-Ríos, S.~Tabik, J.~Luengo, A.~Shihavuddin, and F.~Herrera, ``Coral
  species identification with texture or structure images using a two-level
  classifier based on convolutional neural networks,'' \emph{Knowledge-Based
  Systems}, vol. 184, p. 104891, Nov 2019. [Online]. Available:
  \url{https://linkinghub.elsevier.com/retrieve/pii/S0950705119303569}
\BIBentrySTDinterwordspacing

\bibitem{9254002}
\BIBentryALTinterwordspacing
S.~Tabik, A.~Gomez-Rios, J.~L. Martin-Rodriguez, I.~Sevillano-Garcia,
  M.~Rey-Area, D.~Charte, E.~Guirado, J.~L. Suarez, J.~Luengo, M.~A.
  Valero-Gonzalez, P.~Garcia-Villanova, E.~Olmedo-Sanchez, and F.~Herrera,
  ``Covidgr dataset and covid-sdnet methodology for predicting covid-19 based
  on chest x-ray images,'' \emph{IEEE Journal of Biomedical and Health
  Informatics}, vol.~24, no.~12, pp. 3595--3605, Dec 2020. [Online]. Available:
  \url{https://ieeexplore.ieee.org/document/9254002/}
\BIBentrySTDinterwordspacing

\bibitem{krizhevsky2009learning}
A.~Krizhevsky, G.~Hinton \emph{et~al.}, ``Learning multiple layers of features
  from tiny images,'' 2009.

\bibitem{he2016deep}
K.~He, X.~Zhang, S.~Ren, and J.~Sun, ``Deep residual learning for image
  recognition,'' in \emph{Proceedings of the IEEE conference on computer vision
  and pattern recognition}, 2016, pp. 770--778.

\end{thebibliography}
\bibliographystyle{IEEEtran}

\vfill

\end{document}